\documentclass[a4paper,11pt]{article}
\usepackage[english]{babel}%
\usepackage{hyperref}%
\usepackage{rotating}%
\usepackage{amsmath}%
\usepackage{amssymb}
\usepackage{hyperref}
\usepackage{apacite}
\usepackage{makeidx}
\usepackage{lmodern}      
\usepackage{microtype}
\usepackage{graphicx}
\usepackage{subcaption}
\usepackage{calrsfs}
\DeclareMathAlphabet{\pazocal}{OMS}{zplm}{m}{n}

\usepackage{makeidx}
\usepackage{lscape}
\usepackage{amsmath,amsthm,amssymb}
\usepackage[ansinew]{inputenc}
\usepackage{tikz}
\usetikzlibrary{automata, positioning}
\usetikzlibrary{arrows,positioning,decorations.pathreplacing,shapes}
\usepackage{microtype}
\usepackage{pgfplots}
\usepackage[hmargin=1.5cm,vmargin=1cm]{geometry}
\usepackage{amsmath}
\graphicspath{{Graphics/}}
\definecolor{processblue}{cmyk}{0.96 ,0 ,0 ,0}
\def\bbeta{\mbox{\boldmath $\beta$}}
\def\br{\mbox{\boldmath $r$}}

 \def\bDelta{\mbox{\boldmath $\Delta$}}

 \def\bmu{\mbox{\boldmath $\mu$}}
\def\bGamma{\mbox{\boldmath $\Gamma$}}

\def\bvareps{\mbox{\boldmath $\varepsilon$}}

\def\bLambda{\mbox{\boldmath $\Lambda$}}

\def\bTheta{\mathbf{\Theta}}
\def\bSigma{\mathbf{\Sigma}}

\def\bphi{\mbox{\boldmath $\phi$}}
\def\bb{\mathbf{b}}  \def\bi{\mathbf{i}} 
\def\bE{\mathbf{E}}  \def\bQ{\mathbf{Q}} 
\def\bF{\mathbf{F}} \def\bB{\mathbf{B}} \def\bD{\mathbf{D}} \def\bU{\mathbf{U}}
\def\bV{\mathbf{V}}   
 
 \def\bw{\mathbf{w}}  \def\br{\mathbf{r}}
\def\bu{\mathbf{u}}  \def\by{\mathbf{y}} \def\bY{\mathbf{Y}} \def\b0{\mbox{\bf{0}}}
\def\bs{\mathbf{s}}\def\bS{\mathbf{S}}  \def\bI{\mathbf{I}}
\def\bz{\mathbf{z}}\def\bZ{\mathbf{Z}}   \def\bi{\mathbf{i}}\def\bb{\mathbf{b}}
 \def\bT{\mathbf{T}} \def\bW{\mathbf{W}}  \def\bR{\mathbf{R}}
 \def\bX{\mathbf{X}} \def\bx{\mathbf{x}}
\def\bv{\mathbf{v}} \def\bw{\mathbf{w}}\def\bR{\mathbf{R}}
\def\bR{\mathbf{R}}

\begin{document}
\bibliographystyle{apacite}
\renewcommand{\BBAA}{and}
\renewcommand{\BBAB}{and}
\title{\bf Anomaly Detection in High-Dimensional Bank Account Balances via Robust Methods }
\author{{\bf Federico Maddanu}\thanks{federico.maddanu@gmail.com}\\  University of Rome ``Tor Vergata" \and
{\bf Tommaso Proietti} \\
University of Rome ``Tor Vergata" \and
{\bf Riccardo Crupi} \thanks{The views and opinions expressed are those of the authors and do not necessarily reflect the views of Intesa Sanpaolo, its affiliates or its employees.}\\ 
Intesa Sanpaolo S.p.A.
}

\date{\today}
\maketitle
\begin{abstract}

Detecting point anomalies in bank account balances is essential for financial institutions, as it enables the identification of potential fraud, operational issues, or other irregularities. Robust statistics is  useful for flagging outliers and for providing estimates of the data distribution parameters that are not affected by contaminated observations. However, such a strategy is often less efficient and computationally expensive under high dimensional setting. In this paper, we propose and evaluate empirically several robust approaches that may be computationally efficient in medium and high dimensional datasets, with high breakdown points and low computational time. Our application deals with around 2.6 million daily records of anonymous users' bank account balances.

\vspace{.5cm}

\vspace{.5cm}
\end{abstract}

\section{Introduction \label{ch:intro}}

Real-world datasets are often characterised by
observations that behave differently with respect to the majority of data.
These data points are called \textit{outliers} in statistics and may be generated by incorrect measurements, or they could be also caused by exceptional events, or directly belong to another population. With the growing demand and diverse applications across various fields (such as risk management, cybersecurity, health science, engineering, economics, finance) detecting such anomalies is becoming increasingly crucial.
In particular, detecting outliers in banking time series data is essential for financial institutions, as it enables the identification of potential fraud, operational issues, or other irregularities that could impact performance and risk exposure.

In multivariate time series analysis, anomaly detection can be attempted using diagnostic methods based on classical fitting techniques. However, traditional approaches are highly sensitive to outliers, such that the resulting fitted model fails to identify the deviating points (\textit{masking effect}) or, on the other hand, non-contaminated observations may be incorrectly detected as outliers, determining false positive results (phenomenon referred to as \textit{swamping}).
The approach of robust statistics, introduced by \shortciteA{Huber1964} aims to mitigate these issues by providing estimates of the data distribution's parameters that are not affected by outlier observations.

The minimum covariance determinant (MCD) estimator \shortcite{Rousseeuw84,Rousseeuw85} is one of the first highly robust methods for providing estimates of multivariate location and scatter that are resistant to outlying observations. Denoting by $n$ the total number of observations in the dataset, the MCD approach computes the mean and covariance matrix of the
$h<n$ observations whose covariance matrix has the smallest possible determinant. However, the estimator is suitable for low-dimensional datasets. Specifically, denoting $d$ as the number of time series, we require at least
$d<h$; otherwise, the algorithm will produce a singular covariance matrix, with zero determinant.
\shortciteA{Hubertetal22005} suggest a robust principal component analysis (PCA) approach as a possible solution. They first reduce dimensionality by projecting the observations into a subspace of small to moderate dimension via the first $q$ principal components, and then apply the MCD methodology to obtain robust estimates of the location and scatter parameters.
Unfortunately, whenever $q<d$ the robust PCA approach yields a singular covariance matrix, since it will have $d-q$ eigenvalues equal to zero \shortcite{Hubert18}.

The minimum regularized covariance determinant (MRCD) estimator, proposed by \shortciteA{Boudt20}, was developed to address this issue, providing a robust and nonsingular estimate of the scatter matrix even when $d>n$. The MRCD algorithm operates similarly to the MCD but incorporates a properly regularized covariance matrix in its computations.
A further estimator that performs well in high-dimensional settings was proposed by \shortciteA{Maronna2002} and is based on a modified version of the Gnanadesikan-Kettenring robust covariance estimate \shortcite{Gnanadesikan72}. Named the orthogonalized Gnanadesikan-Kettenring (OGK) estimator, it provides positive-definite and approximately affine-equivariant\footnote{Formally, an estimator \( \hat{\theta} \) is equivariant under a transformation \( g \) if
\[
\hat{\theta}(g(X)) = g(\hat{\theta}(X)).
\]
Examples include location equivariance (e.g., the mean shifts with the data) and affine equivariance (e.g., the covariance matrix transforms accordingly under linear transformations).
} robust scatter matrices, derived from the pairwise robust correlation matrices between each of the $d$ time series.

Further studies on anomaly detection in high-dimensional multivariate time series have been proposed, among others, by \shortciteA{Tsay2000}, \shortciteA{Bianco2001}, and \shortciteA{Pena2023}.

In this paper, we analyze a dataset extracted from the Intesa Sanpaolo (ISP) database, containing daily records of anonymous users bank account balances. The data covers the period from April 1, 2021, to March 31, 2023, comprising 2,636,027 time series observed over 730 days.
The respective univariate time series exhibit multiple seasonality  (related to recurring transactions such as salary, rent, and bills) and a trend component (more related to long term decisions on saving and expenses). Furthermore, the data appears to be strongly affected by outlier observations, such as sharp peaks and level shifts in the mean.

Our approach consists of detecting point anomalies through two main steps. First, we apply robust fitting to estimate the parameters of a deterministic trend-plus-cycle component using the Least Trimmed Estimator (LTE) \shortcite{Rousseeuw84}, similarly to the methods proposed by \shortciteA{Rousseeuw19} and \shortciteA{Rousseeuw06}. Second, robust estimators,such as MRCD and OGK are applied to the resulting robust residuals to detect outliers based on the Mahalanobis Distance (MD).

While these estimators effectively provide robust estimates of location and scatter, they are computationally expensive. To address this, we consider an alternative based on the comedian (COM) estimator \shortcite{Sajesh2012}, which is more computationally efficient for medium- to high-dimensional settings, offering a high breakdown point and fast execution.

However, even the COM estimator becomes impractical for very large datasets (e.g., over one million time series), as storing the corresponding covariance matrix can exceed 1 TB of RAM. In such cases, we perform anomaly detection directly on the robust residuals, using forecasting-based approaches that avoid computing high-dimensional covariance matrices. To this end, we propose both linear and nonlinear robust autoregressive models, identifying outliers through the distribution of squared prediction errors.

In summary, we consider two main families of robust methodologies applied to the ISP dataset: the \textit{distance-based} approaches (including the OGK, MRCD, and COM estimators) and the \textit{forecasting-based} approaches. These methods yield consistent results in terms of both the proportion and type of detected outliers.  Moreover, the set of potentially contaminated series is further refined by integrating the anomaly detection results with a clustering analysis.

We believe the proposed methodology offers a valuable contribution to the literature on anomaly detection in high-dimensional settings.

The rest of the paper is structured as follows. Section \ref{meth} introduces the main problem and describes the methodologies. Simulation studies are presented in Section \ref{sim}. Section \ref{data} presents the ISP dataset and section \ref{app} discusses the corresponding application to outlier detection. Section \ref{secClu} integrates previous results with a clustering analysis.
The last section concludes the paper. Further details are relegated in the Appendix.

\section{Methodology}
\label{meth}

Let us consider $d$ time series, each consisting of $n$ observations, and collect them in the $(n \times d)$ matrix $\bX$, where $\bx_i$, for $i=1, \dots, d$, is an $(n \times 1)$ vector corresponding to the $i$-th time series, such that $\bx_i=(x_{1,i}, \dots, x_{n,i})'$.
We model the individual time series according to the  trend plus seasonal model
\begin{equation}
x_{t,i}=\nu_{t,i}(\bbeta_{i,\nu})  + \gamma_{t,i}(\bbeta_{i,\gamma}) + r_{t,i},   \;\;\; i = 1, \ldots, d,
\label{mdltrend}
\end{equation}
where $\nu_{t,i}(\bbeta_{i,\nu})$ and $\gamma_{t,i}(\bbeta_{i,\nu})$ are the trend and cycle components with respective parameters $\bbeta_{i,\nu}$ and $\bbeta_{i,\nu}$, while $r_{t,i}$ is the short-term component, such that
$\br_t = (r_{1,t},r_{2,t}, \ldots, r_{d,t})'$ has a $d$-variate multivariate distribution with zero mean and scatter parameter $\bSigma$.

One can think of $\nu_{t,i}(\bbeta_{i,\nu})+\gamma_{t,i}(\bbeta_{i,\nu})$ as the regular part of the model. To introduce contamination into the model, we assume that $\bX$ is not directly observable, so that we only observe
\begin{equation}
\bY=\bX+\bS,
\label{mdl1}
\end{equation}
where $\bS$ is a sparse matrix containing outliers in some of its $(t,i)$ entries.

For instance, suppose that $\bx_i$ is not contaminated by outliers; then $\bs_{i} = \b0_n$, where $\b0_n$ is a $(n \times 1)$ column vector with all zero entries. Otherwise, if $\bx_i$ is contaminated, then the elements of $\bs_{i}$ for $t=1, \dots, n$ are defined as
\begin{equation}
 s_{t,i} = \delta_i \omega(L)^{-1} I_t(\tau),
\label{eqout}
\end{equation}
where $L$ is the lag operator, such that $L x_t = x_{t-1}$, $\delta_i \in \mathbb{R}$ represents the effect of the outlier, $\omega(L) = 1 - \omega_1 L - \omega_2 L^2 - \dots - \omega_p L^p$ is a $p$-order polynomial, and
\begin{equation}
I_t(\tau) = \begin{cases} 1, & \text{if } t = \tau, \\ 0, & \text{if } t \neq \tau. \end{cases}
\label{eqind}
\end{equation}
is the indicator function $I_t(\tau)$ of an outlier located at time $t=\tau$,  often referred to a pulse dummy;
$\omega(L)^{-1} I_t(\tau)$ defines the anomaly signature, $\omega(L)^{-1}$ representing the dynamic response to the outlier.

By restricting the coefficients in the $\omega(L)$ polynomial, we can define different types of outliers. For instance, by setting all the coefficients to zero, such that $\omega(L) = 1$, we obtain an \textit{additive outlier} (AO), defined as
\begin{equation}
s_{t,i} = \delta_i  I_t(\tau)
\end{equation}
A \textit{level shift outlier} (LSO) can be obtained by integrating an AO, which results in $\omega(L) = 1 - L$, and is defined as
\begin{equation}
s_{t,i} = \delta_i I(t \geq \tau),
\end{equation}
where $I(t \geq \tau)$ is obtained directly from $(1-L)^{-1} I_t(\tau) = (1+L+L^2+\cdots)I_t(\tau)$.
A further definition can be introduced with $ \omega_p<\omega_{p-1}<\cdots < \omega_2<\omega_1<1$. In this case, the effect of the outlier at time $\tau$ gradually decays over the next $p$ observations.

Our approach to detecting these kinds of anomalies is based on two main steps. First, we robustly estimate the parameters of the trend and cyclical components using the LTE method proposed by \shortciteA{Rousseeuw84}.
To achieve this, we assume that equation (\ref{mdltrend}) follows a deterministic pattern, given by
\begin{equation}
\begin{array}{rcl}
\nu_{t,i}(\bbeta_\nu) & = & \sum_{j=0}^v \beta_{\nu,j} t^j \\
\gamma_{t,i}(\bbeta_\gamma) & = & \sum_{j=1}^c \biggr(\beta_{\gamma,j} \cos(\lambda_j t)  + \beta^*_{\gamma,j} \sin(\lambda_j  t) \biggr),
\end{array}
\label{detTrend}
\end{equation}
where $\lambda_j$ are the seasonal frequencies and the dependence on $i$ is omitted to keep the notation simple. Robust estimation is carried out by \emph{least trimmed  estimator},  which is based on minimizing with the sum of the smallest $h$ squared residuals
\begin{equation}
\sum_{j=1}^h \hat{r}^2_{j:n,i},
\end{equation}
where $\hat{r}^2_{1:n,i} \leq \hat{r}^2_{2:n,i} \leq \cdots \leq \hat{r}^2_{n:n,i}$ and $h$ is usually chosen as $ h=\lfloor 0.75n \rfloor$ See \shortciteA{Rousseeuw84}, \shortciteA{Rousseeuw06}, \shortciteA{Rousseeuw19}, and Appendix \ref{A1} for more details.

The LTE residual is
\begin{equation}
\hat{r}_{t,i}= y_{t,i} -  \sum_{j=0}^v t^j\hat{\beta}_{\nu,j}  - \sum_{j=1}^c \biggr(\hat{\beta}_{\gamma,j} \cos(\lambda_j t)  + \hat{\beta}^*_{\gamma,j} \sin(\lambda_j  t) \biggr)
\label{eqres}
\end{equation}

In the second step, conditional on a robust estimates of the scatter parameter of the distribution of $\br_t$, $\hat{\bSigma}$,  we consider    outlying the $i$-th series characterized by a large MD from the estimated distribution center of $\hat{\bR}=(\hat{\br}_1,\cdots,\hat{\br}_d)$. The MD for $\hat{\br}_i = (\hat{r}_{1,i}, \cdots, \hat{r}_{n,i})'$, the robust residuals for the $i$-th series is defined as follows:
\begin{equation}
\mathcal{D}_i(\hat{\br}_i, \hat{\bSigma})=\sqrt{\hat{\br}_i'\hat{\bSigma}^{-1}\hat{\br}_i}.
\end{equation}

Anomaly detection of a particular observation $(t,i)$, also known as a cellwise outlier, can be performed by analyzing the standardized distances $c^{z}_{t,i}$, which correspond to each $(t,i)$ element of $\bR$,
%
%
\begin{equation}
c^{z}_{t,i}= \mbox{z-score}(c_{t,i}),\;\; c_{t,i}= \frac{\hat{r}_{t,i}^2}{\hat{\sigma}^2_{i,i}}, \label{eqMD}
\end{equation}
where $\hat{\sigma}^2_{i,i}$ is the robust estimate of the variance for the $i$-th variable.
An observation is flagged as an outlier if $c^{z}_{t,i} > \kappa$, where $\kappa$ is a predefined threshold.
We consider different approaches that provide robust estimates of the distribution parameters.

\subsection{Anomaly detection via distance-based methods}

\paragraph*{The OGK estimator.} \shortciteA{Maronna2002} introduced the OGK estimator as a general approach for constructing positive definite and approximately affine equivariant robust scatter matrices, starting from a robust measure of bivariate scatter. Their method was specifically applied to the bivariate covariance estimator proposed by \shortciteA{Gnanadesikan72}. The resulting multivariate scatter estimator is computed as follows:


\begin{enumerate}

\item  Let $s(\cdot)$ as robust univariate estimators of scale. Here, $s(\cdot)$ is computed via the tau-scale estimator proposed by \shortciteA{Yohai88}.

\item  Standardize the data by defining $\bz_t = \bB^{-1}\hat{\br}_t$ for $t = 1, \cdots , n$ where $\bB = \mbox{diag}(s(\hat{\br}_1),\cdots, s(\hat{\br}_d))$.

\item  Compute the "pairwise correlation matrix" $\bU$ of the variables of $\bZ = (\bz_1, \cdots, \bz_d)$, with elements
$\bu_{jk}=\frac{1}{4}(s^2(\bz_j+\bz_k) -s^2(\bz_j-\bz_k))$. $\bU$ is symmetric, but not necessarily positive definite.

\item Let $\bE$ be the matrix of eigenvectors of $\bU$, and let $\bV=\bE \bZ$ and $\bLambda=\mbox{diag}(s^2(\bv_1),\cdots,s^2(\bv_d))$.

\item Compute the scatter estimate for $\bZ$ as $\hat{\bSigma}(\bZ)= \bE \bLambda \bE' $.

\item Obtain the final estimates by turning back to $\bR$ via $\hat{\bSigma}(\bR)_{OGK}=\bB \hat{\bSigma}(\bZ) \bB'$ .

\end{enumerate}


\paragraph*{The MRCD estimator.}\shortciteA{Boudt20} introduced the MRCD estimator, which provides a non-singular robust estimate of the scatter matrix in high-dimensional settings. Unlike the MCD  \shortcite{Rousseeuw84}, this method constructs the scatter matrix as a convex combination of a target matrix and the sample covariance matrix of a selected subset, such that estimates are carried out by minimizing:
\begin{equation}
\mbox{det} \lbrace \rho \bT + (1-\rho) \mbox{Cov}(\bR_H),
\rbrace
\end{equation}
where $\mbox{Cov}(\bR_H)$ is the covariance matrix computed on the $H$-subset and $\bT$ is the positive definite target matrix. Different choices for $\bT$ are proposed in \shortciteA{Boudt20}, in this paper we set $\bT$ equals to the identity matrix. 
Finally, the weight assigned to the target matrix ($\rho$) is determined through a data-driven approach (see \citeauthor{Boudt20}, 2020), ensuring that regularization is applied only when necessary. The MRCD estimator is well-defined in any dimension, inherently well-conditioned, and retains the strong robustness properties of the MCD.
Its implementation follows the steps below:

\begin{enumerate}

\item  Compute the standardized observations $\bz_i = \bD_\bR^{-1} \br_i$, where $\bD_\bR$ is a diagonal matrix whose entries are estimates for the univariate scatter. Here, the tau-scale estimator is employed for the scatter.

\item Given $\bT$, perform singular value decomposition $\bT= \bQ \bLambda \bQ'$ and compute $\bw_i = \bLambda^{-0.5} \bQ'\bz_i$ . 

\item According to \shortciteA{hubert2012deterministic} subsection 3.1, compute 6 robust initial scatter estimates $\bS_j$ for $j = 1, \cdots , 6$.

\item For $j = 1, \cdots , 6$, determine the subsets $H_j$ of $\bW$ containing the $h=0.75n$ observations with lowest MD in terms of $\bS_j$.
\item Let $H^*$ be the subset for which the regularized matrix $\rho \bT + (1-\rho)c_\alpha \bS_W(H_j)$ has the lowest determinant among the candidate subsets, where $c_\alpha  = \alpha^{-1}\chi^2_{\text{cdf}}\left( \chi^2_{\text{inv}}(\alpha, d), d + 2 \right)$ is a consistency factor depending on $\alpha=(n-h)/n$ and $ \bS_W(H) = (h-1)^{-1} \bW'\bW $.

\item Given $H^*$, compute the final robust estimate of $\bSigma$ via $$
\begin{array}{rcl}
\hat{\bSigma}_{MRCD} &=&  \bD_\bR \bQ \bLambda^{1/2}  \biggr(
            \rho \bI + c_\alpha(1-\rho) \bS_W(H^*)                                 \biggr) \bLambda^{1/2} \bQ \bD_\bR
\end{array}
$$
where $$
\begin{array}{rcl}
\bS_W(H^*) &=& (1- h)^{-1} \bZ_{H^*}'\bZ_{H^*}
\end{array}
$$
\end{enumerate}


\paragraph*{The COM estimator.} It turns out that the OGK and MRCD approaches are particularly expensive in terms of computational time. In contrast, the COM estimator \shortcite{Sajesh2012} is computationally efficient, with a high breakdown value and low computation time. This approach has been found to be suitable for anomaly detection, as it can detect a large number of outliers in high-dimensional datasets. It is based on the comedian statistic introduced by \shortciteA{Falk88}
\begin{equation}
 \mbox{COM}(\hat{\br}_j,\hat{\br}_i)=\mbox{med} \biggr ( (\hat{\br}_j - \mbox{med}(\hat{\br}_j))  \odot (\hat{\br}_i - \mbox{med}(\hat{\br}_i))  \biggr),
\end{equation}
for $i,j=1,\cdots,d$, where $\mbox{med}(\cdot)$ is the median and the operator $\odot$ represents the element-by-element product. 
Similarly, we can define the correlation median as \begin{equation}
 \bDelta(\hat{\bR}) = \bD \mbox{COM}(\hat{\bR}) \bD'
\end{equation}
where $\bD=\mbox{diag}(\mbox{MAD}(\hat{\br}_i)^{-1}) $ for $i=1,\cdots,d$ and $\mbox{MAD}(\hat{\br}_i)=\mbox{med}( \vert \hat{\br}_i - \mbox{med}(\hat{\br}_i) \vert)$ is the \textit{median absolute deviation}.


Given that the comedian matrix $\mbox{COM}$, as a robust alternative to the covariance matrix, is generally not positive (semi)-definite, the following steps are necessary:

\begin{enumerate}

\item Compute the eigenvector $\bE$ of the matrix $\bDelta$.

\item  Let $\bQ= \bD^{-1} \bE$ and $\bZ'= \bQ^{-1} \hat{\bR}'$.

\item The resulting robust estimates of location and scatter are
$$
\begin{array}{rcl}
\hat{\bSigma}(\hat{\bR})_{COM} &=& \bQ \bGamma \bQ'
\end{array}
$$
where
$$
\begin{array}{rcl}
\bGamma  &= & \mbox{diag}(\mbox{MAD}^2(\bz_1),\cdots,\mbox{MAD}^2(\bz_d))
\end{array}
$$

\end{enumerate}
The estimates can be improved by replacing $\bDelta$ with $\hat{\bSigma}(\hat{\bR})_{\text{COM}}$ and iterating through the three steps.


\paragraph*{Statistical Features Extraction (FEAU) estimator.} As a benchmark, we also propose a non-robust location and scatter estimator, the FEAU estimator, which is applied directly to the raw data $\bY$ without removing the trend and cyclical components. It is based on the extraction of $p$ statistical features from the data through the following steps:

\begin{enumerate}

\item  Consider $p$ statistical features of the $d$ time series in $\bY$ (as the mean, the standard deviation, the skewness, the kurtosis, the median, the interquartile range etc.), and organize them into the $(p \times d)$ matrix $\bF$.

\item  Standardize $\bF$ as $\bF_z$.

\item Finally, obtains the robust estimates $\hat{\bmu}_{FEAU}$ and $\hat{\bSigma}_{FEAU} $, by computing the mean and covariance  of $\bF_{z}$.

\end{enumerate}

\subsection{Anomaly detection via robust forecasting-based methods}
\label{resAn}

When dealing with very large datasets, computing the MD in equation (\ref{eqMD}) becomes infeasible for standard computers, as it involves a computational complexity of $O(d^2)$, which requires a memory capacity that becomes unmanageable when $d$ is very large. Therefore, applying the distance-based robust methods described above to estimate the scatter is no longer an efficient strategy, and forecasting-based techniques performed directly on the the robust residuals $\hat{r}_{t,i}$ may be preferred. In the sequel, we will drop reference to the $i$-th series in the notation. 

\paragraph*{Robust Heterogeneous Autoregressive (RobHAR) method.} As a first approach, we consider a robust version of the Heterogeneous Autoregressive (HAR) model proposed by \shortciteA{Corsi2009} fitted to $\hat{\bR}$, such that
\begin{equation}
\hat{r}_t = \phi_1 \hat{r}_{t-1} + \phi_7 \bar{r}_{t-7} +   \phi_{30} \bar{r}_{t-30} + \epsilon_t.
\label{AR}
\end{equation}
where $\bar{r}_{t-7} = \frac{1}{7} \sum^7_{i=1} \hat{r}_{t-i}$, $\bar{r}_{t-30} = \frac{1}{30} \sum^{30}_{i=1} \hat{r}_{t-i}$ and the coefficients $\phi_1, \phi_7$ and $\phi_{30}$ are robustly estimated via the LTE, whereas $\epsilon_t \sim WN(0,\sigma^2_{\epsilon}) $.

Point anomalies can be flagged by analyzing the distribution of the resulting squared prediction errors:
\begin{equation}
\hat{\epsilon}^2_t = \left(\hat{r}_t - \hat{\phi}_1 \hat{r}_{t-1} - \hat{\phi}_7 \bar{r}_{t-7} -   \hat{\phi}_{30} \bar{r}_{t-30} \right)^2,
\label{predErrs}
\end{equation}
and checking whether their values exceed a given threshold. The determination of appropriate thresholds will be discussed in subsection \ref{thres}.

The HAR model is widely recognized as a benchmark in financial econometrics due to its ability to effectively capture the long-range persistence in the data. It has also demonstrated good performance and practicality in forecasting realized volatility \shortcite{Proietti2016}.
In our specific case, it is able to capture stochastic seasonal patterns at weekly and monthly frequencies that may not be accounted for by the deterministic component.

\paragraph*{Robust Non-linear Heterogeneous Autoregressive (RobNHAR) method.}
As a further approach, we propose a non-linear extension of the HAR process via the introduction of an univariate neural autoregressive model:
\[
{r}_t = f_\bTheta(\hat{r}_{t-1}, \bar{r}_{t-7}, \bar{r}_{t-30}) + \epsilon_t.
\]
Its architecture is based on a feedforward neural network, a widely used model in time series forecasting applications \shortcite{Zhang98, Qi2008, Suganthan21}, which allows for better capturing possible non-linear structures in the data.

To predict the value at time $t$, the following input vector is constructed:
\[
\bv_t=\begin{bmatrix}
    \hat{r}_{t-1} \\
\bar{r}_{t-7} \\
\bar{r}_{t-30}
\end{bmatrix},
\]
which is passed through a two-layer feedforward neural network composed by:
\begin{enumerate}
    \item \textit{Hidden Layer:} Computes a non-linear transformation via:
    \[
    \mathbf{h}_t = \text{max}(\b0_\ell, \bW_1 \bv_t + \bb_1),
    \]
    where $\bW_1 \in \mathbb{R}^{\ell \times 3}$ is the weight matrix, $\mathbf{b}_1 \in \mathbb{R}^\ell$ is the bias vector, and $\ell$ is the number of hidden units (in our specific case we set $\ell = 10$).
    \item \textit{Output Layer:} Produces the final prediction via the linear transformation:
    \[
    \hat{r}_t = \bw_2' \mathbf{h}_t + b_2,
    \]
    where $\bw_2 \in \mathbb{R}^{\ell}$ and $b_2 \in \mathbb{R}$.
\end{enumerate}
More details on the above neural network architecture can be found in \shortciteA{Zhang21}, Chapter 4.

Robustness to outliers is achieved by training the model using the trimmed mean squared error as the loss function, which focuses on the $h$ smallest residuals:
\begin{equation}
\mathcal{L}(\bW_1, \bw_2, \bb_1, b_2) = \frac{1}{h} \sum_{j=1}^{h} \hat{\epsilon}^2_{i_j},
\label{triMSE}
\end{equation}
where $\hat{\epsilon}^2_{i_1} \leq \cdots \leq \hat{\epsilon}^2_{i_h}$ are the ordered squared residuals, such that the parameters in $\bTheta = (\bW_1, \bW_2, \bb_1, b_2) $ are estimated by minimizing equation (\ref{triMSE}).
As before, point anomalies are detected by analyzing the distribution of the squared prediction errors.

\subsection{Identification of outliers typology}
\label{cusum}

Once outliers have been detected, it is often useful to investigate their statistical nature. In the context of bank account balances, for instance, identifying negative LSOs can be particularly relevant, as they may indicate large withdrawals and alert financial institutions to potentially suspicious activity. Similarly, AOs may also be linked to malicious operations, for example, when a significant amount of money is deposited into an account and withdrawn on the same day.

The distance-based methods proposed in the previous sections are generally effective at distinguishing between AOs and LSOs by performing anomaly detection directly on the raw robust residuals or on their first differences, respectively, although some overlap may occur. In contrast, the RobHAR and RobNHAR methodologies do not offer the same level of discrimination, and therefore an additional strategy is required to separate the two types of outliers.

In this regard, we propose a heuristic approach to disentangle such anomalies, inspired by the CUSUM statistic introduced by \shortciteA{Page54}. Let $\hat{r}^{z}_t$ denote the standardized robust residuals, and suppose an outlier is detected at time $t = \tau$. We compute two local statistics, $S_b$ and $S_a$, defined as the averages of $\hat{r}^{z}_t$ over the intervals $\left[ \tau - w_0, \tau - 1 \right]$ and $\left[ \tau + 1, \tau + w_0\right] $, respectively, where $w_0$ is an appropriate window size.\footnote{In the empirical application to ISP data, we set $w_0 = 30$.}

Given $\hat{\sigma}_r$ as the standard deviation of $\hat{r}^{z}_t$ (which is approximately 1, since residuals are standardized), we classify the anomaly as a LSO if $\vert S_b - S_a \vert > 2\hat{\sigma}_r$. Alternatively, if both $\vert \hat{r}^{z}_\tau - S_a \vert > 2\hat{\sigma}_r$ and $\vert \hat{r}^{z}_\tau - S_b \vert > 2\hat{\sigma}_r$, we classify it as an AO. If neither condition is satisfied, no classification is assigned.

Once an outlier is detected, its sign can be inferred by introducing a dummy regressor into equation (\ref{eqres}) and recomputing the robust residuals. In the case of an LSO, we include the dummy variable $I_t(t > \tau)$, while for AOs, we consider $I_t(t = \tau)$. The sign of the estimated coefficient associated with the dummy then indicates whether the outlier represents a positive or negative deviation from the expected level.

\section{Simulation study}
\label{sim}

This section presents the results of a Monte Carlo experiment aimed at assessing the finite sample properties of the robust methodologies previously described. In particular, we first consider the non-robust FEAU approach and the three robust covariance estimators, all of which are directly applied to the raw data. Then, the distance and forecasting based methods are applied to the residuals obtained after removing the trend and cyclical components estimated via LTE. To represent the forecasting based approaches, we consider a simple robust autoregressive process of order 1 (RobAR(1)). Specifically, we restrict the model to be linear and set $\phi_7 = \phi_{30} = 0$, since the simulated series does not exhibit non-linearity and the seasonal pattern is assumed to be fully captured by the deterministic component.

We conduct 500 replications.
In each replication, we simulate the $(d \times n)$ matrix $\bY = \bX + \bS$, where $d = 600$ and $n = 400$.
Contamination is introduced in 40\% of the dataset, such that outliers appear in the first $240$ series at observation $\tau = 80$.

We consider both AO and LSO contamination, with outlier effects
\(\delta_i = (\tilde{\sigma}_{i,i}, 1.5 \tilde{\sigma}_{i,i})\), where \(\tilde{\sigma}_{i,i}\) is the sample standard deviation of the \(i\)-th series.

In general, an observation is flagged as an outlier if the standardized MD distances in equation (\ref{eqMD}) exceed a given threshold \(\kappa = Q(\cdot)\), where \(Q(\cdot)\) is the quantile function of the empirical distribution of the standardized distances. If we consider the RobAR(1) approach, outlier detection is performed by analyzing the distribution of the squared prediction errors in equation (\ref{predErrs}).


\subsection{The data generating process (DGP)}
\label{DGP}

The DGP for the variable $\bX$ follows a structural time series model:
\begin{equation}
    x_{it} = \upsilon_{it} + \gamma_{it} + \varepsilon_{it},
\end{equation}
where $\upsilon_{it}$ represents the local level trend component:
\begin{equation}
\begin{array}{rcl}
    \upsilon_{it} & = & \upsilon_{i,t-1} + \beta_{i,t-1} + \epsilon_{it}, \quad \epsilon_{it} \sim N(0,\sigma^2_\epsilon), \\
    \beta_{it} & = & \beta_{i,t-1} + \zeta_{it}, \quad \zeta_{it} \sim N(0,\sigma^2_\zeta).
\end{array}
\end{equation}

The cyclical component $\gamma_{it}$ is given by:
\begin{equation}
    \gamma_{it} = s_{it}(j) + \eta_{it}, \quad \eta_{it} \sim N(0,\sigma^2_\eta),
\end{equation}
where $s_{it}(j)$ is modeled as a monthly seasonal cubic polynomial:
\begin{equation}
    s_{it}(j) = a_{ij} + b_{ij} (t-t_j) + c_{ij}(t-t_j)^2 + d_{ij}(t-t_j)^3, \quad \text{if} \quad t \leq t_j.
\end{equation}
Here, $t_j$ are predefined control points representing monthly seasonality, ensuring that the periodic condition holds:
$
    s_{it}(j) = s_{i(t+30)}(j)
$.
For the sake of simplicity, this experiment does not account for a weekly cycle.
Finally, the noise term $\varepsilon_{it}$ follows a multivariate normal distribution, s.t.
$
    \bvareps_i \sim N(\mathbf{0}_n, \bSigma_\varepsilon).
$.
The DGP is simulated with the following parameter values:
\[
\begin{array}{ccccccccc}
    \sigma^2_\epsilon & \sigma^2_\zeta & \sigma^2_\eta & \sigma_s & a_{i1} & a_{i2} & a_{i3} & a_{i4} & \delta_i \\
    \hline
    \hline
    10^{-3}\varphi_1  & 10^{-3}\varphi_2 & 0   & 10^2\varphi_3 & \sigma_s \varphi_4  & \sigma_s \varphi_4  & \sigma_s \varphi_5  & \sigma_s \varphi_5  & \hat{\sigma}_{i,i}
\end{array}
\]
with control points set as $t_1=0, t_2=14, t_3=15$, and $t_4=30$. The parameters $\varphi_q$ ($q=1,\dots,5$) are drawn from a uniform distribution $\mathcal{U}(0,1)$ to introduce variability in the simulation.

\subsection{Monte Carlo results}
\label{MC}

Tables \ref{tabAO} and \ref{tabLSO} present the Monte Carlo results in terms of the percentage of identified outliers and the number of false positives. In the first column, the output is obtained using a standard threshold of \( Q(0.9975) \), while in the second column, the threshold is optimized to minimize the number of false positives (less than 0.01\% of false positives generated by the robust estimators applied to the residuals of equation (\ref{eqres})).

Results on AOs detection are reported in Table \ref{tabAO}. The approach based on the two-step procedure described in the previous section, demonstrates strong performance. When outliers are more pronounced (\(\delta_i = 1.5\tilde{\sigma}_{i,i}\)), our method detects about 77\% of AOs with fewer than 10 false positives when using the optimized threshold. Conversely, when outliers are less evident (\(\delta_i = \tilde{\sigma}_{i,i}\)), it identifies approximately 63\% of AOs while generating around 15 false positives.

In comparison, robust estimators applied directly to the raw data \( \bY \) detect approximately 23\% of outliers in the most evident case and about 2-3\% in the least evident case. In all scenarios, robust approaches outperform the non-robust estimator FEAU.

Similar results hold for LSOs detection, as shown in Table \ref{tabLSO}. Our approach identifies 66\% of LSOs with around 7 false positives under the highest threshold and strongest outlier effect. Interestingly, the non-robust estimator performs similarly to the robust estimators applied directly to the raw data, and their performance is not significantly different from the two-step methodology. When outliers are less pronounced, approximately 46\% of LSOs are detected using the two step procedure, and the advantage over the other methods becomes more evident.

\begin{figure}
\begin{center}
\includegraphics[width=1.10\linewidth]{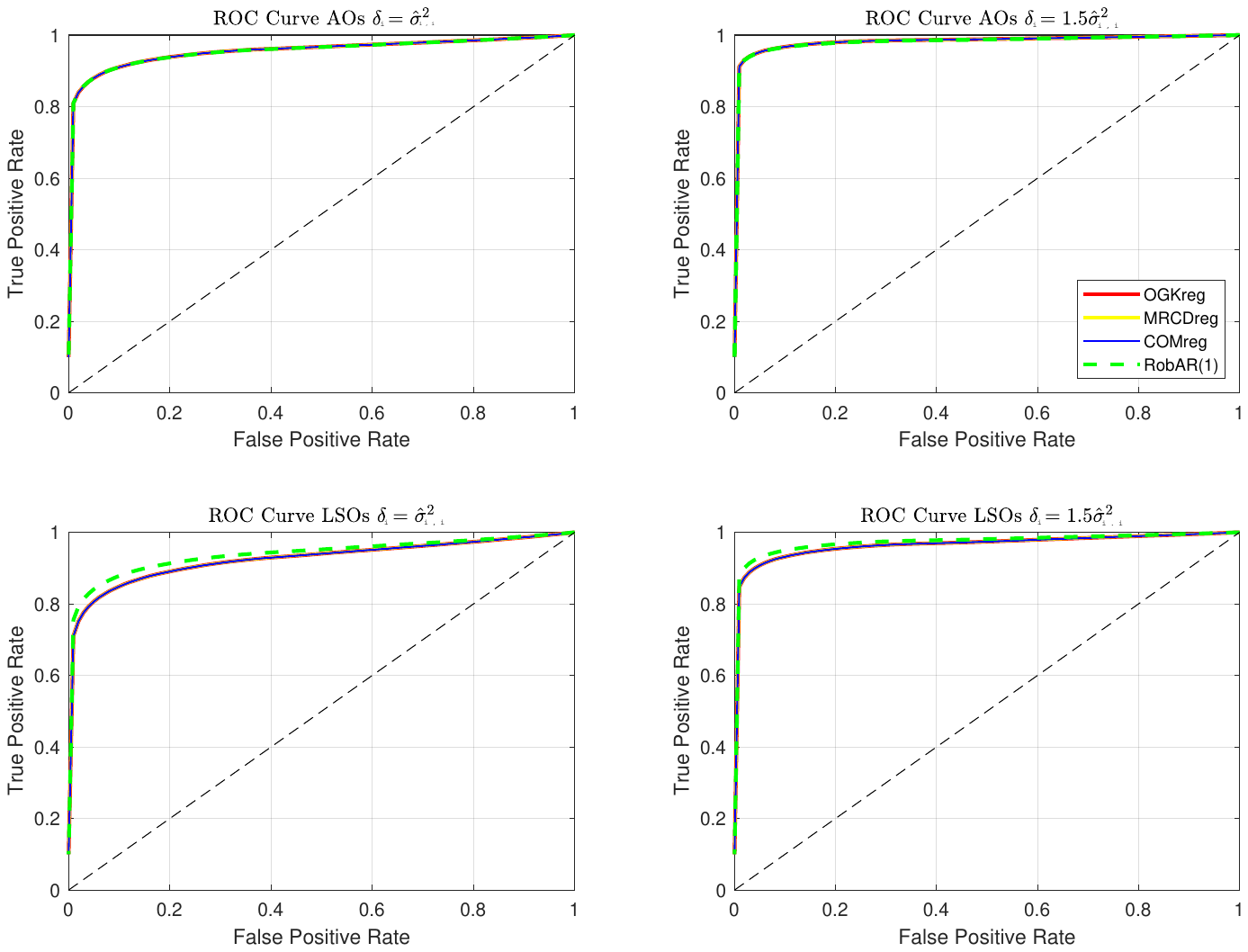}
\end{center}
\caption{\footnotesize ROC curve for the OGK, MRCD, COM and RobAR(1) methodologies applied to the robust residuals in detecting AOs and LSOs with outlier effect $\delta_i = \lbrace \hat{\sigma}^2_{i,i}, 1.5 \hat{\sigma}^2_{i,i} \rbrace$.}
\label{ROC}
\end{figure}

The performance of the OGK, MRCD, COM and RobAR(1) approaches applied to the robust residuals is further illustrated by the receiver operating characteristic (ROC) curve in Figure \ref{ROC}, which plots the true positive rate (i.e., the number of correctly detected outliers over the total number of actual outliers) against the false positive rate (i.e., the number of false positives over the total number of non-contaminated observations) at each threshold setting $Q(k)$, for $k$ varying in $(0,1)$. The different methodologies exhibit very similar performance in detecting AOs, while the forecasting-based method performs slightly better in detecting LSOs.

To summarize, the experiment clearly shows that robust estimators cannot be directly applied to the raw data. If trend or cyclical patterns are present, capturing these components robustly is essential. Additionally, the approach based on the RobAR(1) process and the COM estimator performs almost identically to the two-step alternatives, suggesting a preference for these methods due to their lower computational cost.


\begin{table}
\centering
\begin{tabular}{l||rr|rr}
$\delta_i=\tilde{\sigma}_{i,i}$ & \multicolumn{2}{c}{$\kappa =Q(0.9975)$} & \multicolumn{2}{c}{$\kappa =Q(0.9993)$} \\
\hline
\hline
method   & percOut  &  numFalsPos  & percOut  &  numFalsPos   \\
\hline
     OGK     &     0.1098    & 573.64 &     0.0242 &    162.19  \\ 
     MRCD    &     0.0979 &    576.50  &    0.0217 &    162.80  \\
     COM     &    0.1200 &    571.20 &     0.0308 &    160.61  \\
     FEAU    &     0.0402 &    590.36 &     0.0221 &    162.69  \\
     OGKreg  &    0.7444 &    421.34 &     0.6345 &      15.73 \\ 
     MRCDreg &     0.7453 &    421.13 &     0.6352 &    15.55   \\ 
     COMreg  &   0.7445 &    421.33  &    0.6342 &    15.79   \\ 
     RobAR(1)   &   0.7451 &    419.18  &    0.6339 &    15.85  \\       
\end{tabular}
\begin{tabular}{l||rr|rr}
\hline
\hline
$\delta_i=1.5\tilde{\sigma}_{i,i}$ & \multicolumn{2}{c}{$\kappa =Q(0.9975)$} & \multicolumn{2}{c}{$\kappa =Q(0.9992)$} \\
\hline
\hline
method   & percOut  &  numFalsPos  & percOut  &  numFalsPos   \\
\hline
     OGK      &   0.3128  &     524.92  &    0.2428  &     133.73    \\
     MRCD     &    0.3102  &     525.55   &    0.2389  &     134.67   \\
     COM      &   0.3080  &     526.08   &   0.2368   &    135.17    \\
     FEAU     &    0.0886  &    578.73    &  0.0643    &  176.56     \\
     OGKreg   &     0.8698  &      391.25  &    0.7765  &     5.63   \\
     MRCDreg  &      0.8705  &     391.09   &    0.7769  &     5.54    \\
     COMreg   &     0.8698  &     391.24   &   0.7763   &     5.69     \\
     RobAR(1)    &  0.8712  &     388.90   &   0.7774   &     5.42    \\
     
\end{tabular}
\caption{\footnotesize Monte Carlo results are reported in terms of the percentage of detected AOs (percOut) and the number of false positives (numFalsPos), considering different outlier effects ($\delta_i$) and thresholds ($\kappa $). The methods OGK, MRCD, and COM represent three robust estimators applied directly to the raw data $\bY$. OGKreg, MRCDreg, and COMreg refer to the same estimators applied to the residuals in equation (\ref{eqres}), obtained by removing the trend and cyclical components from $\bY$ using LTE. FEAU concerns the non robust method applied to the raw data. In the RobAR(1) fitting, AOs are detected if the squared prediction error exceed a given threshold.}
\label{tabAO}
\end{table}

\begin{table}
\centering
\begin{tabular}{l||rr|rr}
$\delta_i=  \tilde{\sigma}_{i,i}$ & \multicolumn{2}{c}{$\kappa =Q(0.9975)$} & \multicolumn{2}{c}{$\kappa =Q(0.9995)$} \\
\hline
\hline
method   & percOut  &  numFalsPos  & percOut  &  numFalsPos   \\
\hline
     OGK      &      0.5283   &   471.22  &    0.3568  &    34.37   \\
     MRCD     & 0.5378    &  468.94   &   0.3747   &   30.07    \\
     COM      &   0.5301    &  470.78   &   0.3520   &   35.51    \\
     FEAU     &  0.5398    &  468.44   &   0.3766   &   29.62   \\
     OGKreg   &     0.6281    &  447.25   &   0.4639   &   8.67    \\
     MRCDreg  &   0.6287    &  447.10   &     0.4646 &     8.49    \\
     COMreg   &  0.6276    &  447.37   &    0.4642  &    8.60   \\
   RobAR(1)   &  0.6770    &    434.52 &      0.4824&      3.22   \\   
\end{tabular}
\begin{tabular}{l||rr|rr}
\hline
\hline
$\delta_i=1.5\tilde{\sigma}_{i,i}$ & \multicolumn{2}{c}{$\kappa =Q(0.9975)$} & \multicolumn{2}{c}{$\kappa =Q(0.9993)$} \\
\hline
\hline
method   & percOut  &  numFalsPos  & percOut  &  numFalsPos   \\
\hline
     OGK       &   0.7678  &    413.72  &     0.6437 &     13.51  \\
     MRCD      &    0.7690  &    413.45  &    0.6469  &     12.74    \\
     COM       &   0.7693  &    413.37  &    0.6451  &    13.18  \\
     FEAU      &    0.7707  &    413.03  &    0.6480  &      12.48   \\
     OGKreg    &    0.7884  &    408.78  &    0.6670  &    7.93    \\
     MRCDreg   &    0.7886  &    408.73  &    0.6673  &    7.85  \\
     COMreg    &   0.7880  &    408.89  &    0.6670  &    7.93    \\
     RobAR(1)    &  0.8236  &    399.34  &    0.6827  &    3.15    \\
\end{tabular}
\caption{\footnotesize Monte Carlo results are reported in terms of the percentage of detected LSOs (percOut) and the number of false positives (numFalsPos), considering different outlier effects ($\delta_i$) and thresholds ($\kappa $). The methods OGK, MRCD, and COM represent three robust estimators applied directly to the first difference of the raw data $\bY$. OGKreg, MRCDreg, and COMreg refer to the same estimators applied to the residuals in equation (\ref{eqres}), obtained by removing the trend and cyclical components from $(1-L)\bY$ using LTE. FEAU concerns the non robust method applied to the first difference of the raw data. Finally, via the RobAR(1) fitting LSOs have been detected on the first difference of the robust residuals.}
\label{tabLSO}
\end{table}

\subsection{Threshold selection}
\label{thres}

In the previous section, the thresholds in the second column of Tables \ref{tabAO} and \ref{tabLSO} were set by minimizing the number of false positives, considering the trade-off between false positives and the percentage of detected outliers.
Obviously, this approach is not feasible when dealing with real data, as we cannot define a false positive \textit{a priori}. Considering the distance-based methods, we may focus on the logarithmic tail of the standardized MD histogram. Since natural distributions typically exhibit a consistent pattern in their tails, significant deviations may indicate anomalous values. Therefore, we fit a linear regression to the log-transformed tail, setting the threshold at the point where the data significantly deviates from this linear fit.

\begin{figure}
\begin{center}
\includegraphics[width=1.10\linewidth]{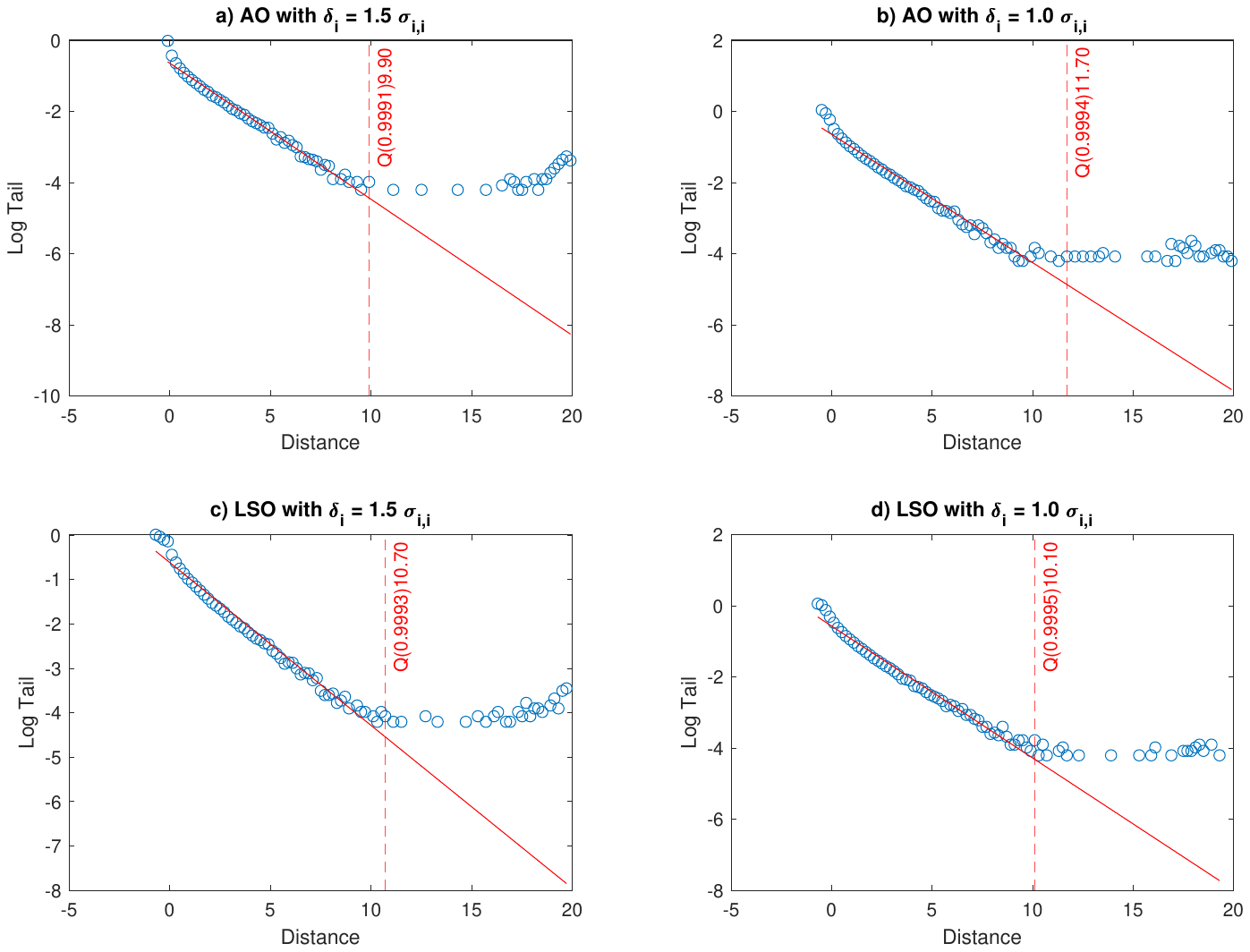}
\end{center}
\caption{\footnotesize Thresholds and linear fit on the log tails of the standardized MD distribution obtained via the COM estimator by considering the DGP in subsection \ref{DGP} under the same assumptions of the Monte Carlo experiment in subsection \ref{MC}, for a) AO $\delta_i= 1.5 \sigma_{i,i}$, b) AO $\delta_i= \sigma_{i,i}$, c) LSO $\delta_i= 1.5 \sigma_{i,i}$ and d) LSO $\delta_i= \sigma_{i,i}$.}
\label{FigThre}
\end{figure}

As an illustrative example, Figure \ref{FigThre} shows the linear fits on the log tails of the standardized MD distributions, obtained via the COM estimator under the same assumptions of the Monte Carlo experiment in sections \ref{DGP} and \ref{MC}, for both AOs and LSOs with  $\delta_i=\lbrace 1, 1.5 \rbrace \hat{\sigma}_{i,i}$. Notice that the resulting thresholds are quite close to the optimal ones reported in Tables \ref{tabAO} and \ref{tabLSO}.

Similarly, this approach can be directly applied to the log distribution of the mean squared errors in equation (\ref{predErrs}), in order to set appropriate thresholds in the forecasting-based methodologies.

When working with real-world data, a linear fit may not always be the most appropriate choice for determining the threshold, and alternative approaches can be considered. In particular, for the ISP datasets, we apply a Pareto fit to the log-tail distribution of the standardized MD (or the standardized prediction errors for the forecasting-based methods) computed from the first differences of the robust residuals, according to $f(z)=\frac{z_m^{\varrho}}{z^{\varrho+1}}\varrho $ with $z \geq z_m$ for each real $z$, while we maintain the linear fit for the non-differenced series.

\section{The ISP dataset}
\label{data}

The ISP dataset contains the daily records of anonymous users' bank account balances, collected over 730 days for a total amount of 2,636,027 of time series. Due to recurring transactions such as salary payments, rent, and bills, as well as varying spending patterns across weekdays, the data exhibit both monthly and weekly seasonality. Additionally, long-term decisions on saving and expenses may generate trend patterns.

\begin{figure}
\begin{center}
\includegraphics[width=1.0\linewidth]{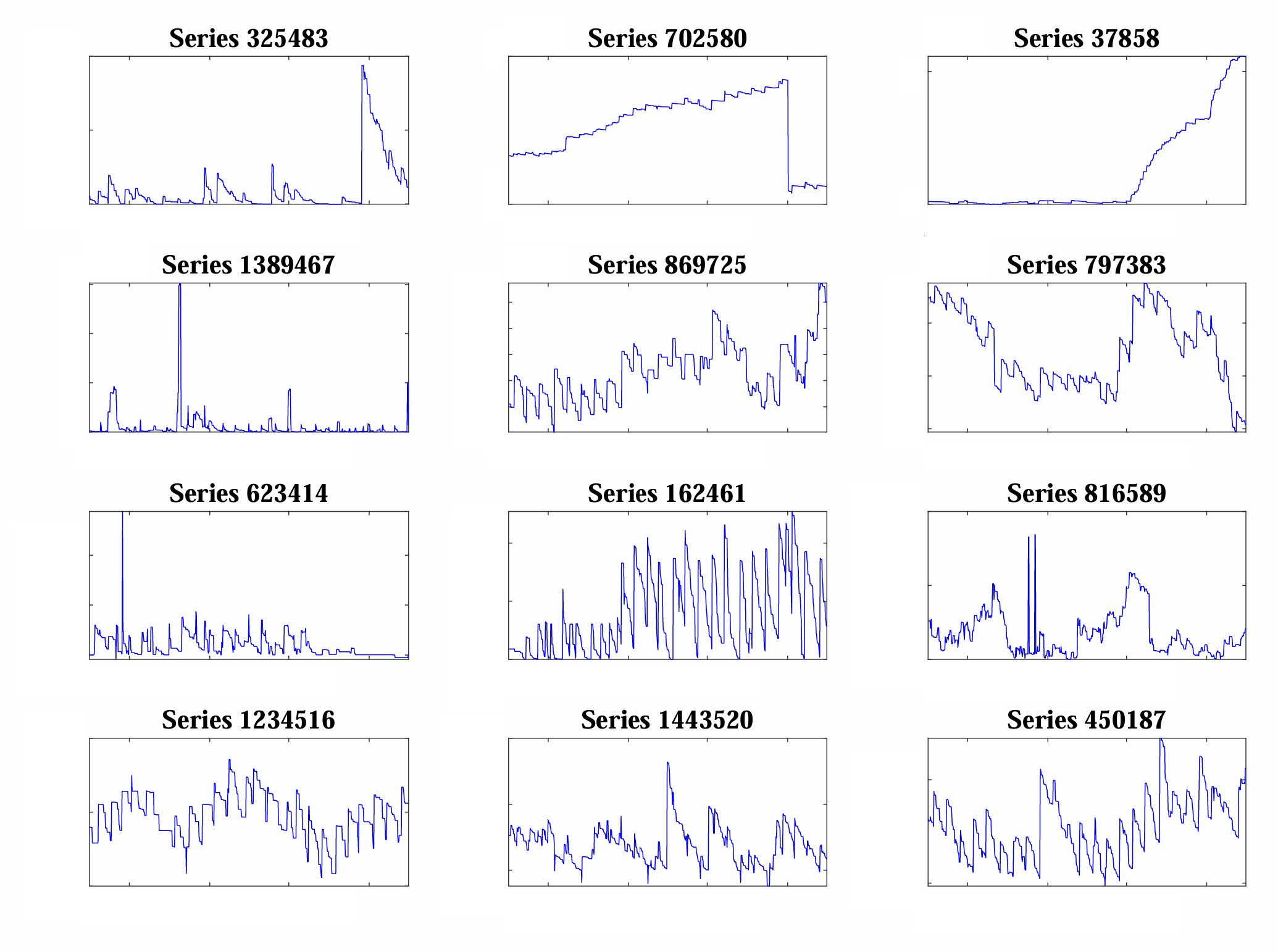}
\end{center}
\caption{\footnotesize 12 bank account balances time series randomly extracted from the main dataset. }
\label{series}
\end{figure}

Figure \ref{series} shows some illustrative examples of these time series, randomly extracted from the main dataset.
Notice that the data are also characterized by recurring sharp peaks and level shifts.
Such anomalies may be due to unexpected expenses or money inflows (e.g., repair the car or receiving an inheritance), but they could also result from potential fraud, operational issues, or other irregularities.
Detecting such outliers is our main purpose, regardless of their benign or malicious nature.

In order to test the robust methods with the higher computational cost, we will first consider two sub-samples consisting of 5,000 and 50,000 bank account balances randomly extracted from the main dataset, which we denote as $\bY_{5K}$ and $\bY_{50K}$.

Before moving on with the analysis, we pre-process the data so that series with fewer than 103 operations (i.e., at least 1 operation per week on average) are excluded from the sample. Moreover, we also exclude dormant accounts with no transactions for at least 200 consecutive days, as well as accounts that remained inactive after May 1st, 2021, or were closed before March 1st, 2023.

As a result, the main dataset $\bY$ consists of 1,503,880 series, while the sub-samples $\bY_{5K}$ and $\bY_{50K}$ contains 2,820 and 28,425 series, respectively.

\begin{figure}
\begin{center}
\includegraphics[width=1.10\linewidth]{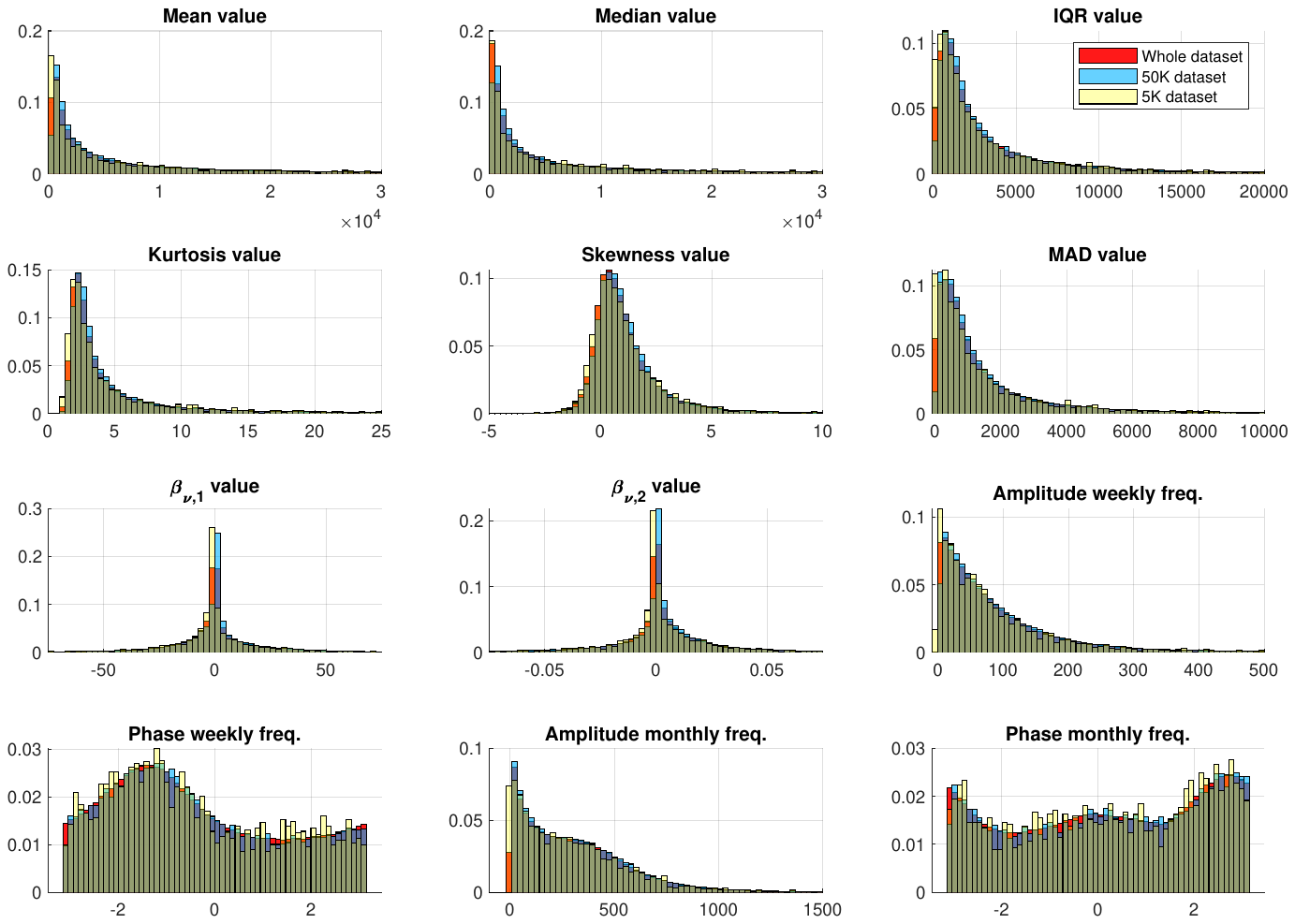}
\end{center}
\caption{\footnotesize Histograms of the mean, median, standard deviation, MAD, kurtosis, skewness, $\beta_{\nu,1}$, $\beta_{\nu,2}$, amplitudes, and phases for the first two harmonics. The latter are computed using the parameters in Equation \ref{mdltrend}. }
\label{kern}
\end{figure}

To verify that the two sub-samples are representative of the entire dataset $\bY$, we examine the distribution of several statistical features extracted from the two sets. Figure \ref{kern} shows the histograms of the mean, median, standard deviation, MAD, kurtosis, skewness, and the parameters of the trend-plus-cycle components (the latter estimated via LTE). Notice that the distributions exhibit fairly similar patterns, indicating that $\bY_{5K}$ and $\bY_{50K}$ are representative of the entire dataset.


\section{Outliers detection on the ISP dataset	}
\label{app}

In the following, we implement point anomaly detection using the two-step procedure described in section \ref{meth} on the reduced datasets $\bY_{5K}$ and $\bY_{50K}$, as well as on the full dataset $\bY$.

The first step is identical for all datasets: robust residuals are computed by removing a deterministic trend and cyclical component from the raw data via the LTE. Based on our empirical experience with these data, a quadratic trend combined with the weekly and monthly cycles (s.t. $\nu=c=2, \lambda_1=2\pi/7$ and $\lambda_2=2\pi/30$ in equation (\ref{detTrend})) is sufficient to capture the persistent structure typically observed in the series.

The second step varies depending on the dataset. For the $\bY_{5K}$ sub-sample, all the methods discussed above are applied. In the case of $\bY_{50K}$, only the COM estimator and the two forecasting-based procedures are considered, in order to reduce computational burden. Finally, due to the high dimensionality of $\bY$, anomaly detection on the full dataset is performed exclusively using the RobHAR and RobNHAR procedures.




\subsection{Outliers detection on the $\bY_{5K}$ dataset}
\label{app5K}

We begin by performing anomaly detection on the smallest dataset, $\bY_{5K}$, using the two-step procedure described in section \ref{meth}. The OGK, MRCD, and COM estimators are applied to the robust residuals in equation (\ref{eqres}) and to their first differences, in order to better capture level shift outliers (LSOs). Since many anomalies may not be purely additive or level shifts, but rather follow the structure in equation (\ref{eqout}), duplicates may arise between the two sets of results. To address this, potential duplicates are removed, and the two outputs are merged into a single array.

Similarly, the two forecasting-based methodologies are applied to the robust residuals ${\bR}$ and their first difference, according with section \ref{resAn}.

Table \ref{tab5k} reports the results obtained from the five approaches, in terms of the amount of point anomalies detected (NumOtlrs), percentage of contaminated time series (PercAllout), percentage of series with exactly one outlier (Perc1out), with two outliers (Perc2out), and with more than two outliers (Perc3out). Notably, all distance-based methods yield similar outcomes, flagging approximately 26\% of the series as contaminated. On the other hand, the RobHAR and RobNHAR approaches appear to be less sensitive, flagging less point anomalies and no more than one outlier per series. Furthermore, the percentage of commonly identified anomalies across all the methods exceeds approximately 90\%.

\begin{table}
\centering
\resizebox{0.7\textwidth}{!}{
\begin{tabular}{c|ccccc}
   & NumOtlrs & PercAllout & Perc1out  &  Perc2out &  Perc3out   \\
\hline
\hline
OGK  &  815 &   26.49\% &   24.54\% &   1.52\%   & 0.43\% \\
MRCD &  779 &   25.99\% &   24.50\% &   1.35\%   & 0.14\% \\
COM  &  804 &   26.38\% &   24.57\% &   1.52\%   & 0.28\% \\
RobHAR  &  554 &   19.65\% &   19.65\% &   0.00\%   & 0.00\% \\
RobNHAR  &  580 &   20.57\% &   20.57\% &   0.00\%   & 0.00\% \\
\hline
\hline
\end{tabular}}
\caption{\footnotesize Outliers detected in the $\bY_{5K}$ dataset by the OGK, MRCD, COM, RobHAR and RobNHAR methods in terms of total point anomalies detected (NumOtlrs), percentage of contaminated time series (PercAllout), percentage of time series contaminated by just 1 outlier (Perc1out),  percentage of time series contaminated by only 2 outliers (Perc2out) and the percentage of time series contaminated by more than 2 outliers (Perc3out).}
\label{tab5k}
\end{table}

As discussed above, the presence of a large number of contaminated bank account series can largely be attributed to exceptional user activities, where certain transactions are flagged as outliers due to significant deviations from typical user behavior. Therefore, it is of interest to further investigate the statistical nature of these anomalies. For instance, financial institutions may be alerted by widespread, large-scale withdrawals occurring across multiple users within a specific period.

To detect this type of negative LSOs, we further filter the subset of contaminated robust residuals using the methodology described in section \ref{cusum}. This allows us to isolate outlier typologies from the broader set of anomalies reported in Table \ref{tab5k}. The magnitude of each outlier's effect can then be quantified by including the corresponding dummy variables as regressors in equation (\ref{eqres}).

\begin{figure}
\begin{center}
\begin{subfigure}[b]{0.45\textwidth}
\includegraphics[width=1.15\linewidth]{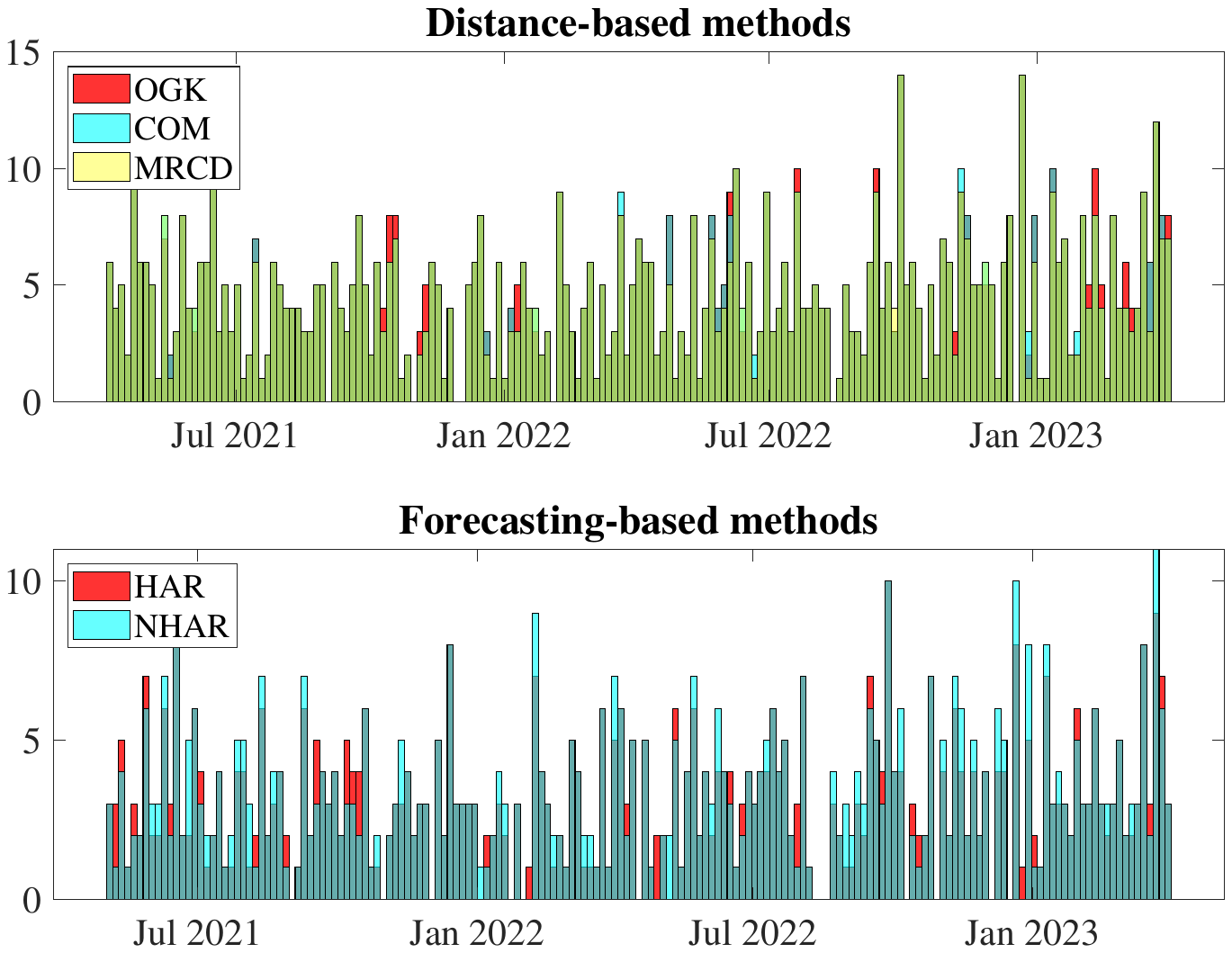}
\caption{\footnotesize Histogram of the total amount of outliers detected via the five methods.}
\end{subfigure}
\begin{subfigure}[b]{0.45\textwidth}
\includegraphics[width=1.15\linewidth]{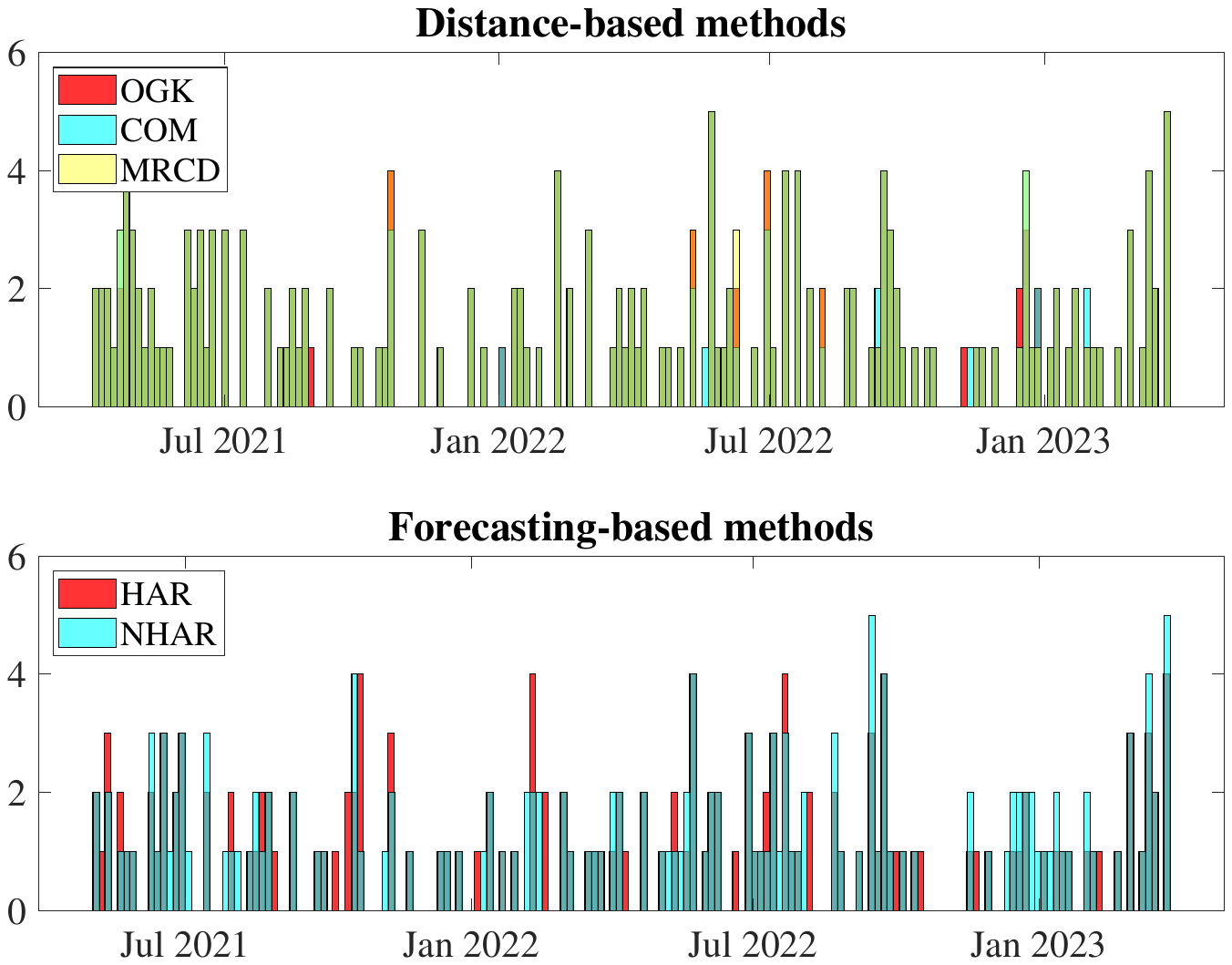}
\caption{\footnotesize Histogram of the negative LSOs detected via the five methods.}
\end{subfigure}
\begin{subfigure}[b]{0.45\textwidth}
\includegraphics[width=1.15\linewidth]{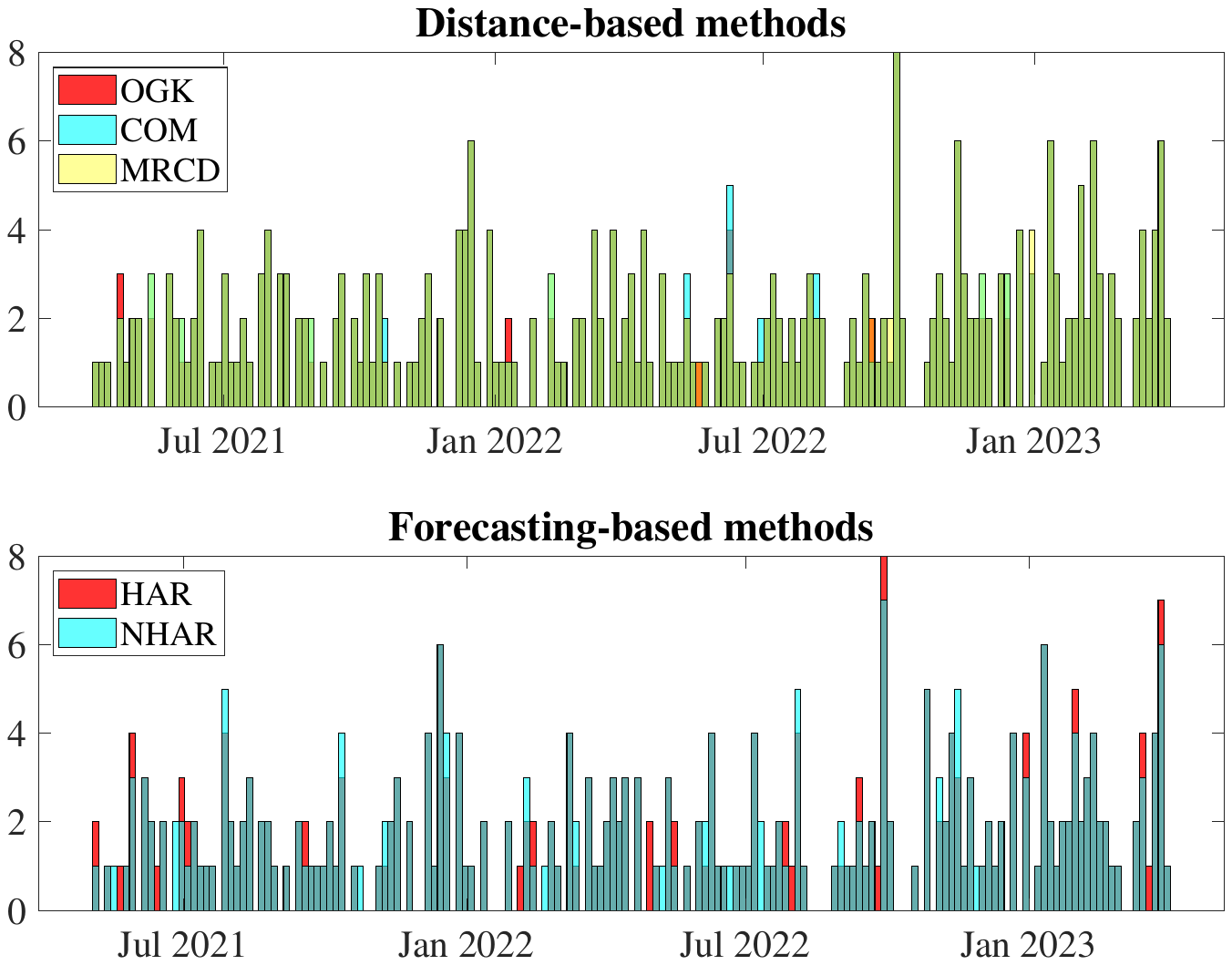}
\caption{\footnotesize Histogram of the positive LSOs detected via the five methods.}
\end{subfigure}
\begin{subfigure}[b]{0.45\textwidth}
\includegraphics[width=1.15\linewidth]{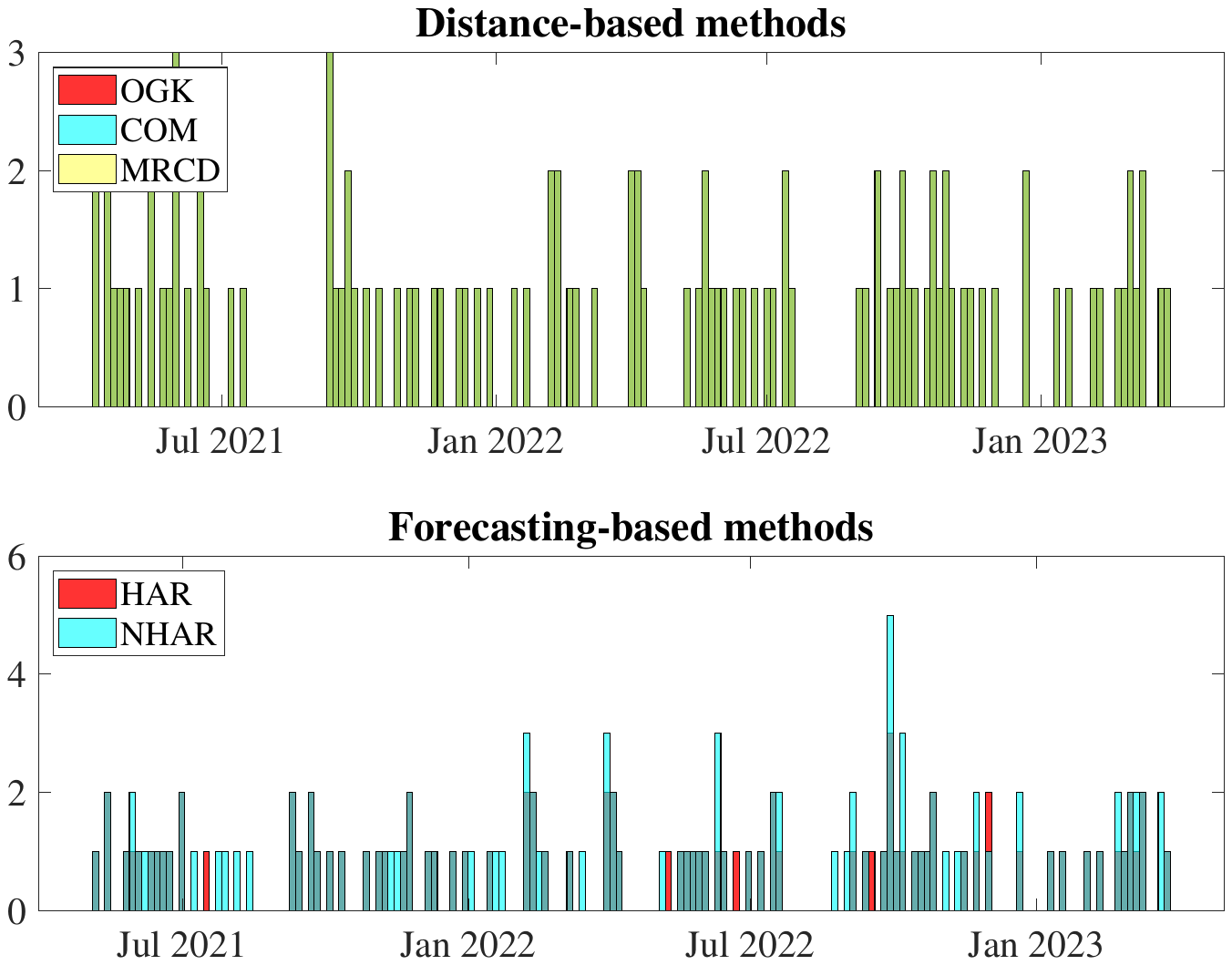}
\caption{\footnotesize Histogram of the total amount of AOs detected via the five methods.}
\end{subfigure}
\end{center}
\caption{\footnotesize Outliers histograms for the five methodologies by considering the reduced $\bY_{5K}$ dataset.}
\label{hist5k}
\end{figure}

Figure \ref{hist5k} displays the distribution of outliers corresponding to: a) the total number of outliers detected by the four methods, b) negative LSOs, c) positive LSOs, and d) AOs.
The various histograms appear to be uniformly distributed over time.

\subsection{Outlier detection on the $\bY_{50K}$ dataset}
\label{app50K}

This subsection presents the results obtained by applying anomaly detection to the $\bY_{50K}$ sub-sample. As before, we begin by computing the robust residuals using the LTE. Given the high computational costs associated with the OGK and MRCD methods, only the more computationally efficient COM estimator and forecasting-based procedures are considered in the second step. 

As an illustrative example, consider a synthetic dataset $\tilde{\bY}$ generated according to equation (\ref{mdltrend}), with 40\% outlier contamination and the data-generating process described in section \ref{DGP}. For $n=730$ and $d=1460$, the computational time required by the OGK and MRCD estimators is 12.87 and 14.32 minutes, respectively, whereas the COM estimator requires only 78.34 seconds.

\begin{figure}
\begin{center}
\begin{subfigure}[b]{0.45\textwidth}
\includegraphics[width=1.15\linewidth]{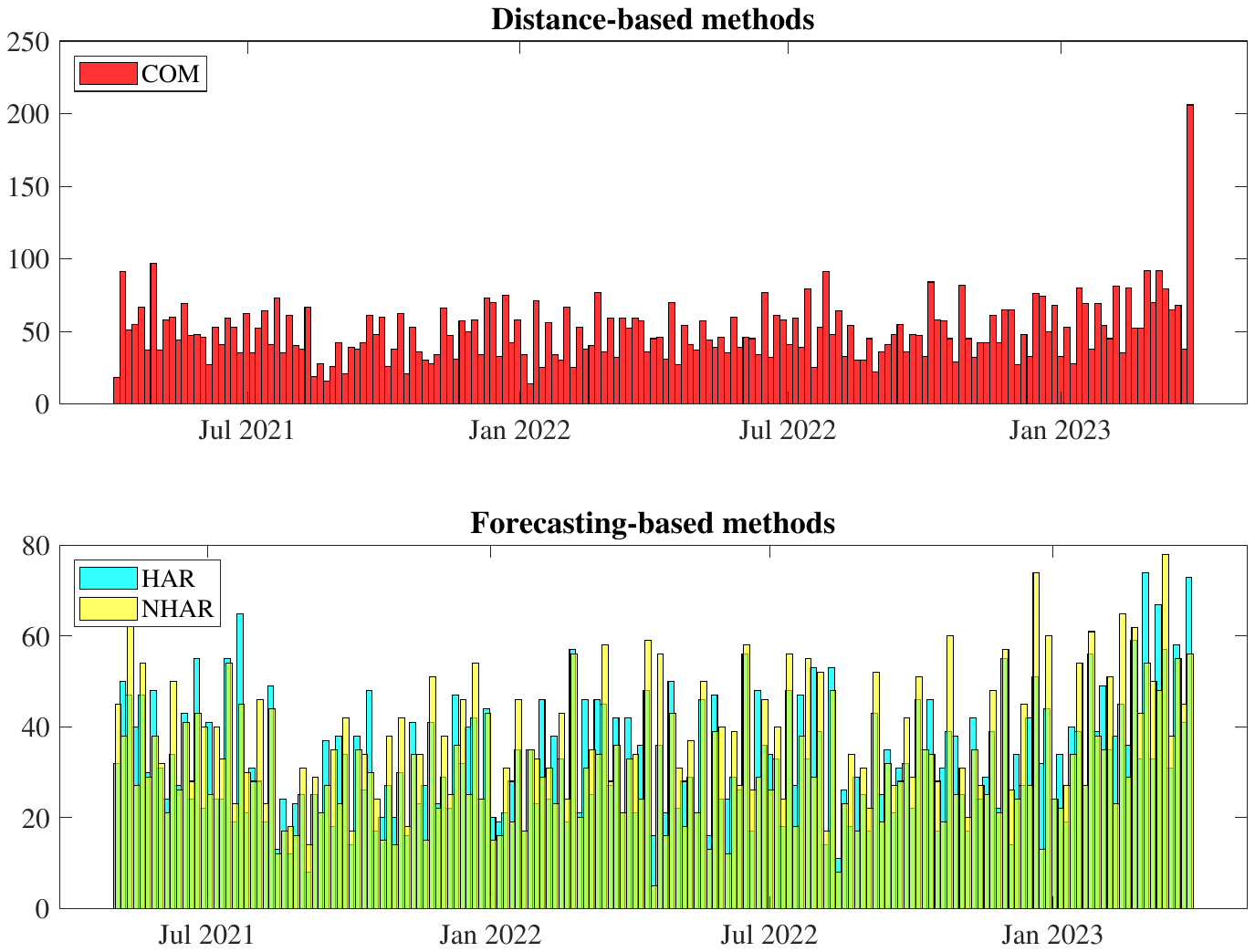}
\caption{\footnotesize Histogram of the total amount of outliers detected via the three methods.}
\end{subfigure}
\begin{subfigure}[b]{0.45\textwidth}
\includegraphics[width=1.15\linewidth]{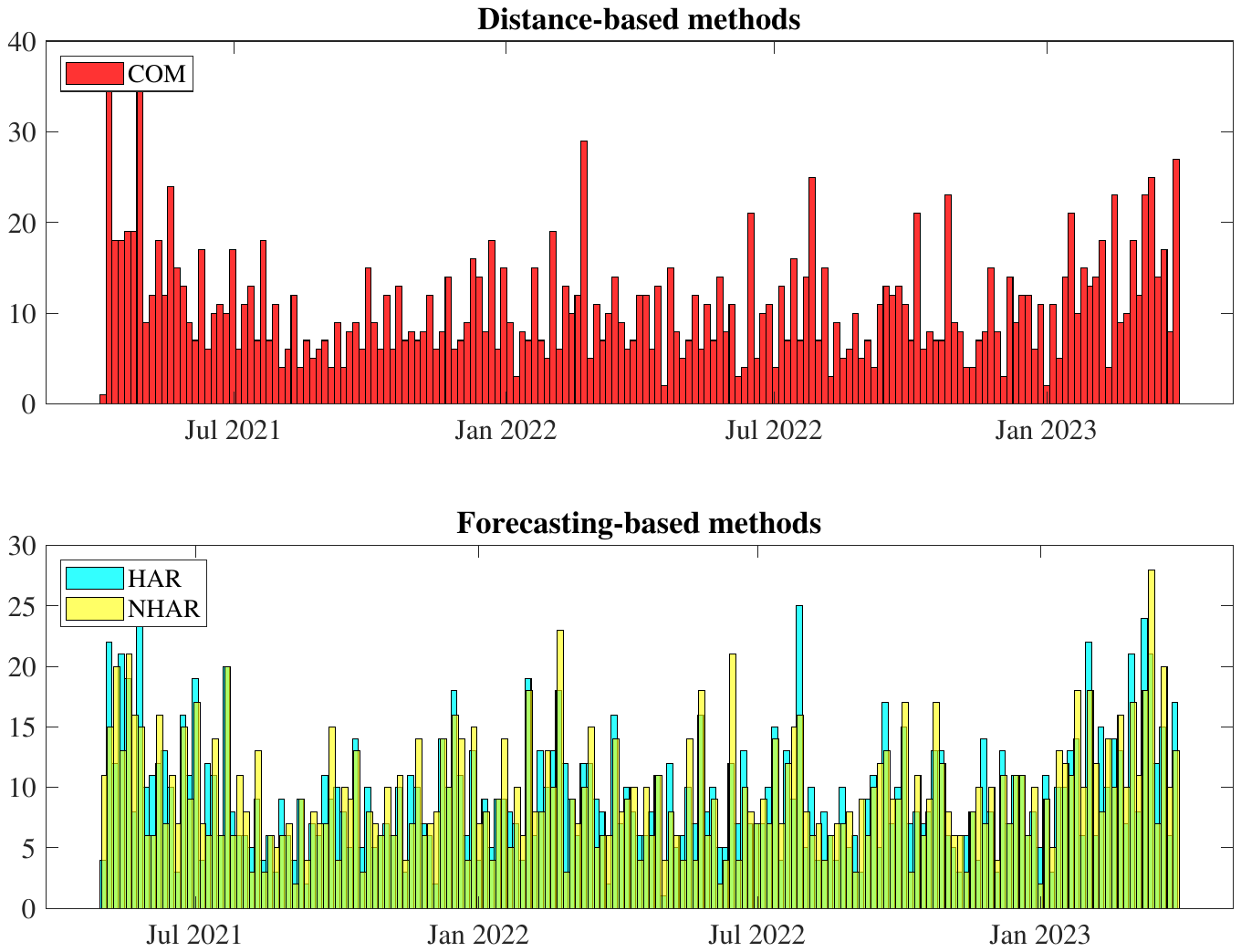}
\caption{\footnotesize Histogram of the negative LSOs detected via the three methods.}
\end{subfigure}
\begin{subfigure}[b]{0.45\textwidth}
\includegraphics[width=1.15\linewidth]{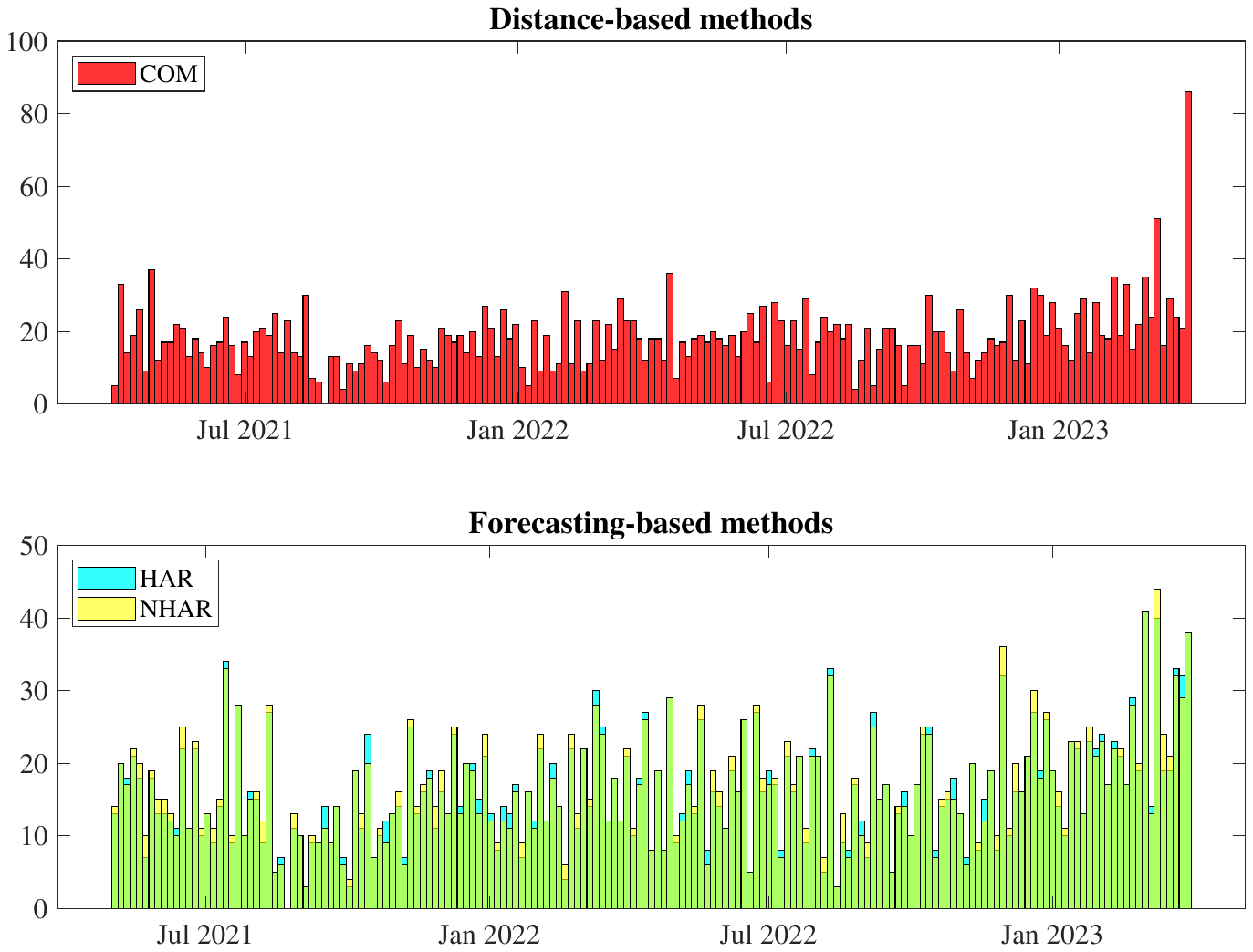}
\caption{\footnotesize Histogram of the positive LSOs detected via the three methods.}
\end{subfigure}
\begin{subfigure}[b]{0.45\textwidth}
\includegraphics[width=1.15\linewidth]{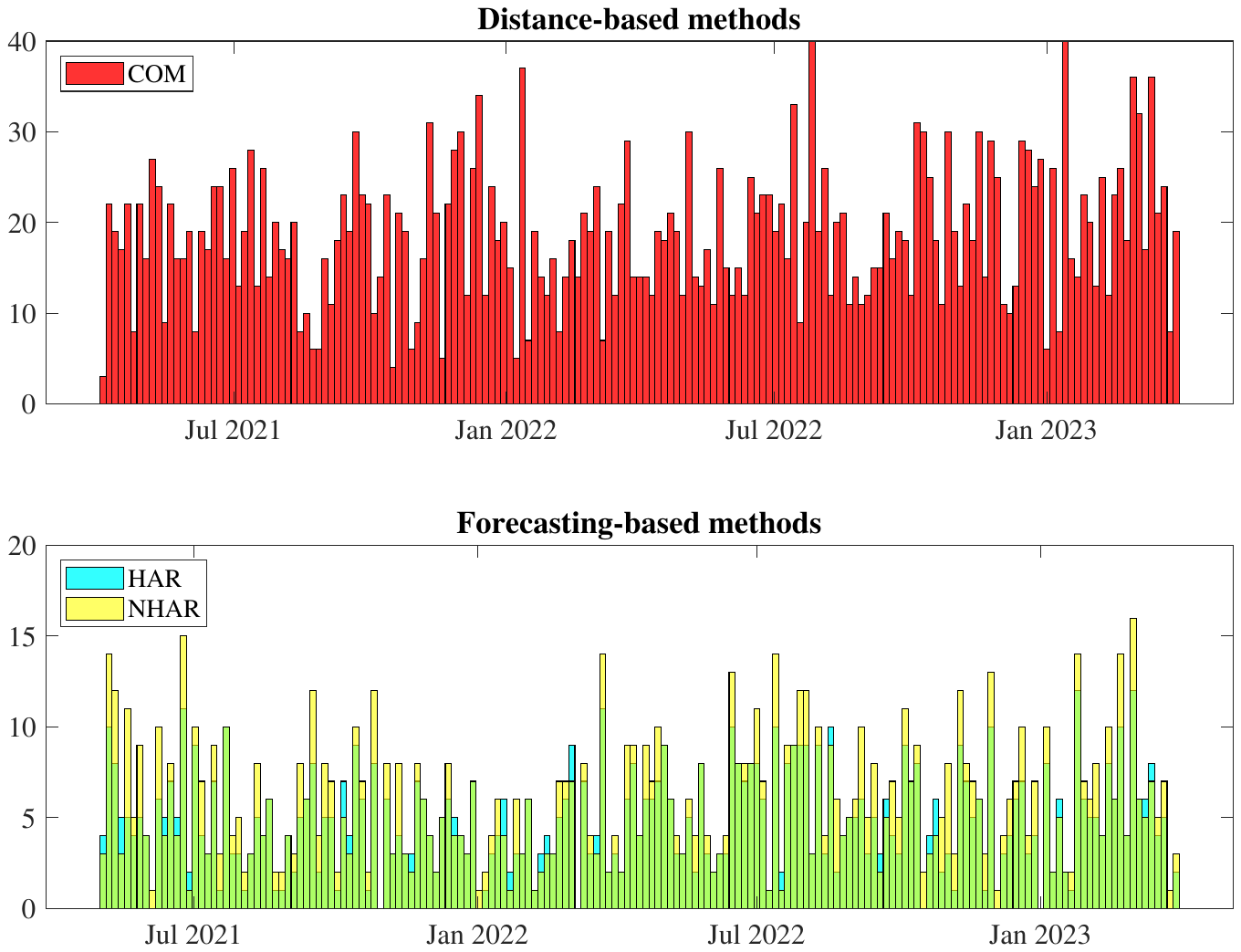}
\caption{\footnotesize Histogram of the AOs detected via the three methods.}
\end{subfigure}
\end{center}
\caption{\footnotesize Outliers histograms for the three methodologies by considering the reduced $\bY_{50K}$ dataset.}
\label{hist50k}
\end{figure}

Table \ref{tab50k} reports the number and percentage of contaminated time series identified by the COM, RobHAR and RobNHAR methods. The proportion of anomalies is almost identical than that observed in the $\bY_{5K}$ dataset, with approximately 27\%, 20\% and 21\% of the series flagged as contaminated by the COM, RobHAR and RobNHAR methodologies, respectively, which share over 90\% of common outliers detected. AOs and LSOs are then detected among the contaminated robust residuals using the heuristic procedure described in section \ref{cusum}. The corresponding histograms are shown in Figure \ref{hist50k}, revealing a distribution pattern broadly similar to that observed in the previous analysis of the $\bY_{5K}$ dataset. However, a noticeable peak at the end of March 2023 appears in the histogram of positive LSOs detected by the COM estimator, an anomaly not captured by the other methods.

   \begin{table}
\centering
\resizebox{0.7\textwidth}{!}{
\begin{tabular}{c|ccccc}
   & NumOtlrs & PercAllout & Perc1out  &  Perc2out &  Perc3out   \\
\hline
\hline
COM  &  8762 &   27.17\% &   24.54\% &   1.72\%   & 0.91\% \\
RobHAR  &  5816 &   20.46\% &     20.46\% &   0.00\%   & 0.00\% \\
RobNHAR &  5984 &   21.00\% &     20.95\% &   0.05\%   & 0.00\%\\ \hline
\hline
\end{tabular}}
\caption{\footnotesize Outliers detected in the $\bY_{50K}$ dataset by the COM, RobHAR and RobNHAR methods in terms of number of total point anomalies detected (NumOtlrs), percentage of contaminated time series (PercAllout), percentage of time series contaminated by just 1 outlier (Perc1out),  percentage of time series contaminated by only 2 outliers (Perc2out) and the percentage of time series contaminated by more than 2 outliers (Perc3out).}
\label{tab50k}
\end{table}

\subsection{Outlier detection on the entire dataset $\bY$}

We now discuss the results of anomaly detection on the entire ISP dataset $\bY$. As in previous cases, we begin by computing the robust residuals, removing the deterministic trend and cyclical components via the LTE.

While the methodology described in Section \ref{meth} could, in principle, be applied to $\bY$, even the most computationally efficient robust estimator COM requires storing a very large covariance matrix ($1.5 \times 1.5$ million entries), making the operation infeasible on standard hardware due to a memory requirement of approximately 1 TB of RAM.

To overcome this limitation, outlier detection is performed using only the forecasting-based methods, which do not rely on the computation of high-dimensional covariance matrices. In particular, to further reduce computational time in the RobNHAR approach, the feedforward network is trained only on the reduced $\bY_{50K}$ sample, while predictions are performed on the entire $\bY$ dataset.

\begin{figure}
\begin{center}
\includegraphics[width=1.00\linewidth]{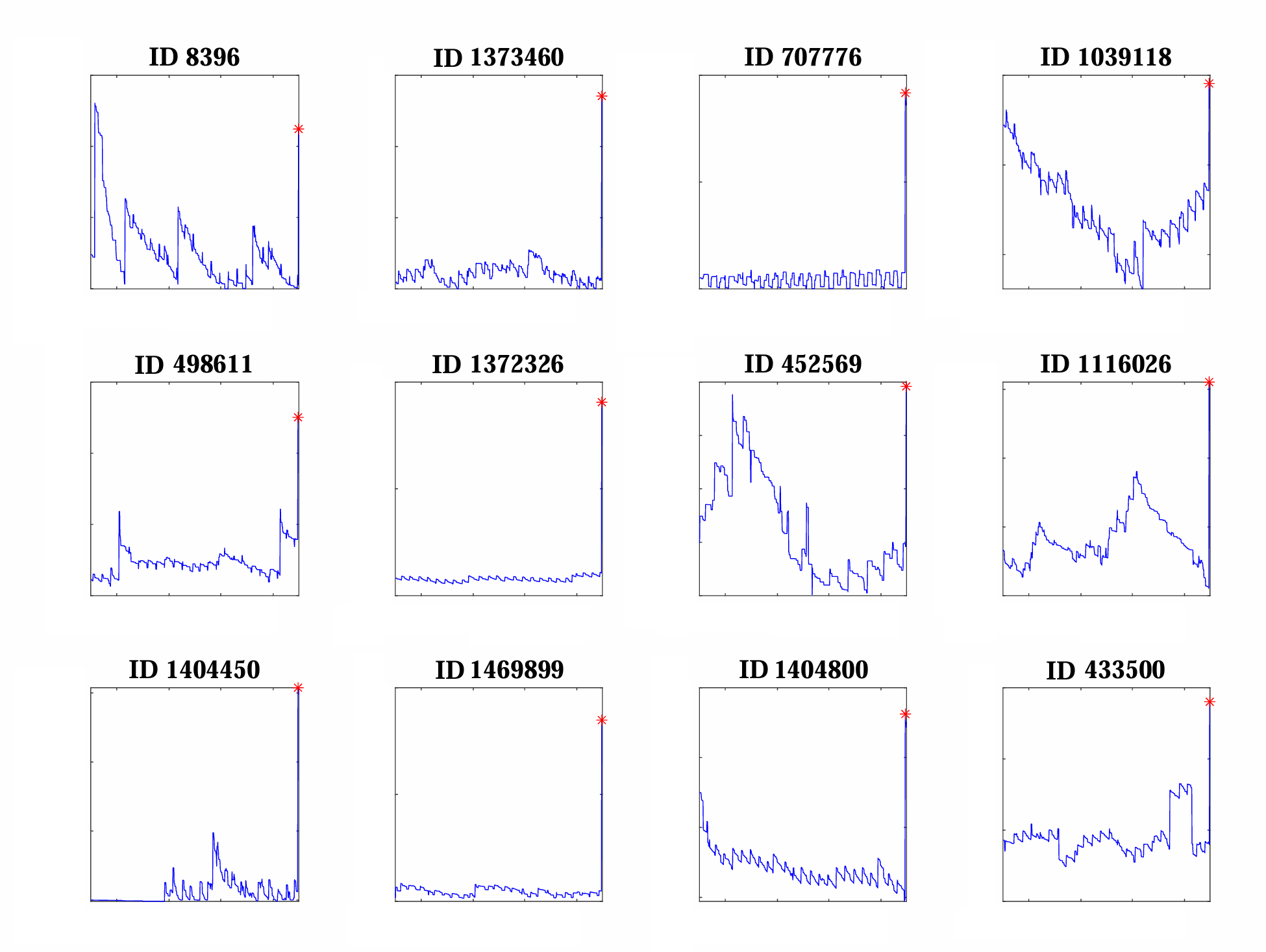}
\end{center}
\caption{\footnotesize Contaminated time series with positive LSOs detected the last week of March 2023 according to section \ref{cusum}. The red asterisks indicate the time positions of the outliers.}
\label{march23}
\end{figure}

The two methods share more than 94\% of common outliers detected and, as shown in Table \ref{tab2M}, identify approximately 20\% and 21\% of the time series as contaminated, corresponding to a total of 301,860 and 322,757 point anomalies, respectively. These findings are consistent with those obtained from the smaller datasets. Similarly, the distribution of AOs and LSOs detected according to Section \ref{cusum} confirms the previously observed patterns, including a notable peak in the histogram of positive LSOs during the last week of March 2023. While we do not have a definitive explanation for this behavior, a sample of such anomalies is highlighted by the red asterisks in Figure \ref{march23}.

\begin{figure}
\begin{center}
\includegraphics[width=1.00\linewidth]{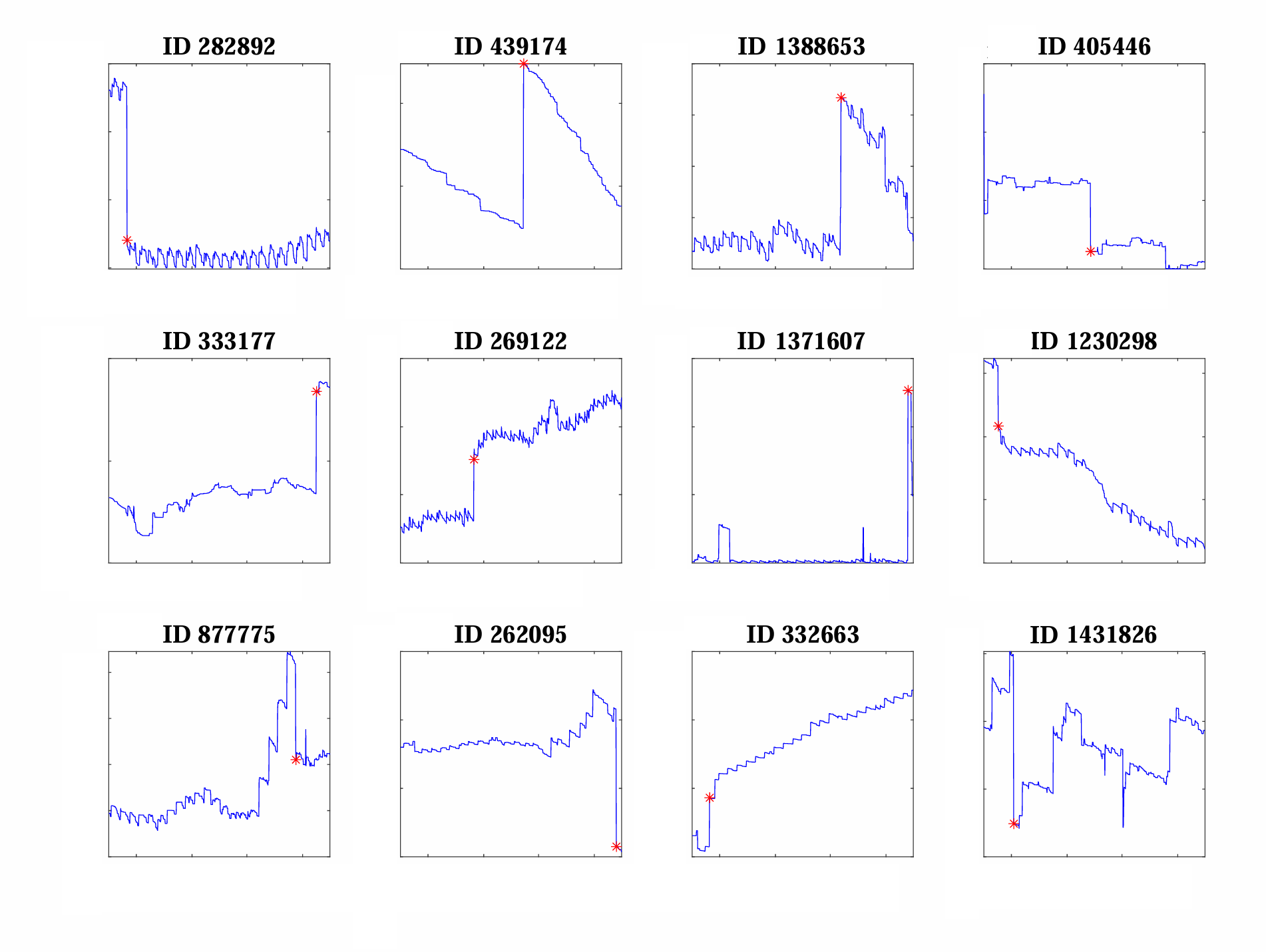}
\end{center}
\caption{\footnotesize Contaminated time series with LSOs detected according to section \ref{cusum}. The red asterisks indicate the time positions of the outliers.}
\label{negOut}
\end{figure}

\begin{figure}
\begin{center}
\includegraphics[width=1.00\linewidth]{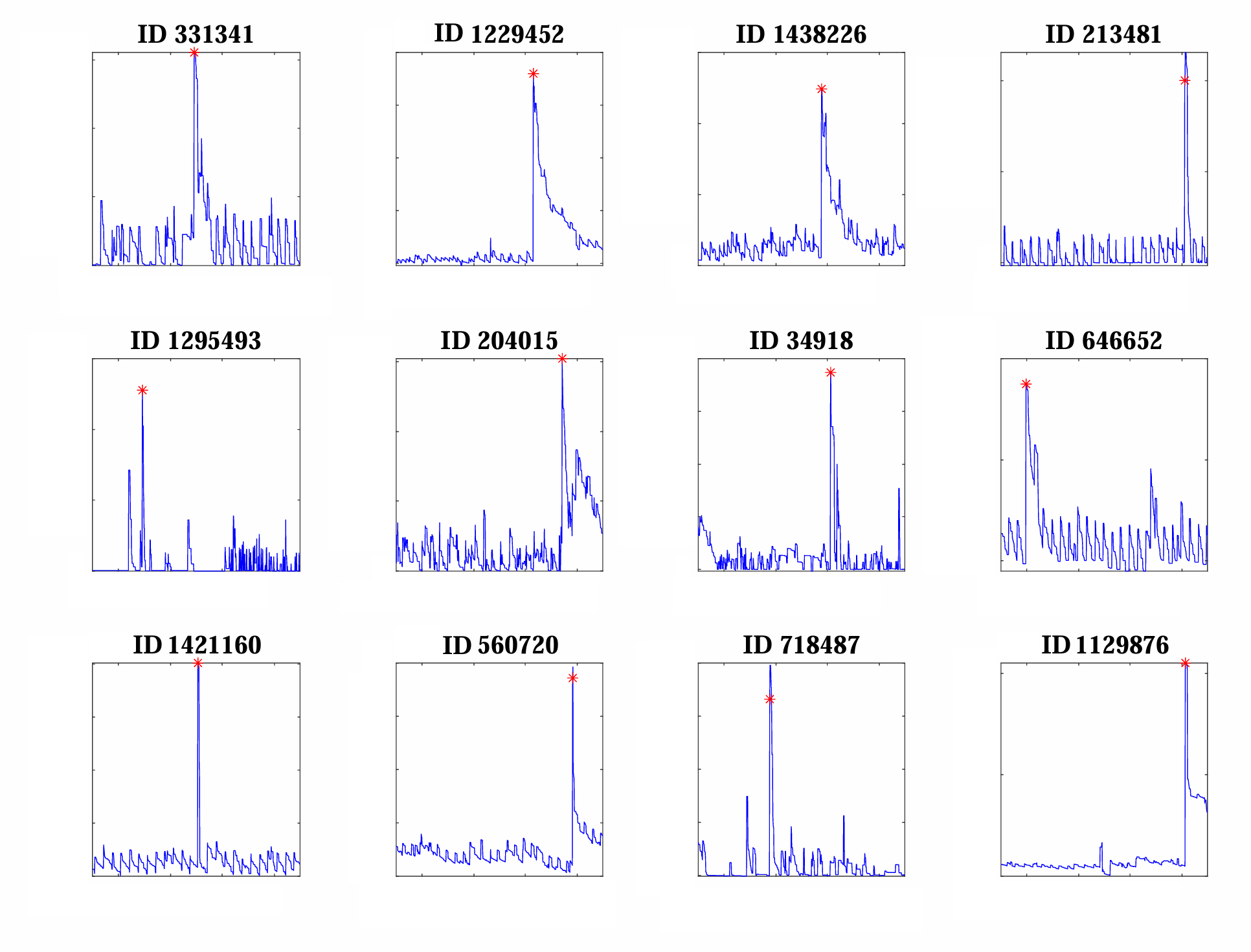}
\end{center}
\caption{\footnotesize Contaminated time series with AOs detected according to section \ref{cusum}. The red asterisks indicate the time positions of the outliers.}
\label{nidfOut}
\end{figure}

Finally, we also present illustrative examples from the main dataset that demonstrate how our approach operates, showing contaminated series with AOs and LSOs, indicated by red asterisks in Figures \ref{negOut} and \ref{nidfOut}, respectively.

   \begin{table}
\centering
\resizebox{0.7\textwidth}{!}{
\begin{tabular}{c|ccccc}
   & NumOtlrs & PercAllout & Perc1out  &  Perc2out &  Perc3out   \\
\hline
\hline
RobHAR  &  301,860 &   20.07\% &     20.07\% &   0.00\%   & 0.00\% \\
RobNHAR &  322,757 &   21.46\% &     21.46\% &   0.00\%   & 0.00\%\\ \hline
\hline
\end{tabular}}
\caption{\footnotesize Outliers detected in the entire sample by the RobHAR and RobNHAR methods in terms of number of total point anomalies detected (NumOtlrs), percentage of contaminated time series (PercAllout), percentage of time series contaminated by just 1 outlier (Perc1out),  percentage of time series contaminated by only 2 outliers (Perc2out) and the percentage of time series contaminated by more than 2 outliers (Perc3out).}
\label{tab2M}
\end{table}

\begin{figure}
\begin{center}
\begin{subfigure}[b]{0.45\textwidth}
\includegraphics[width=1.15\linewidth]{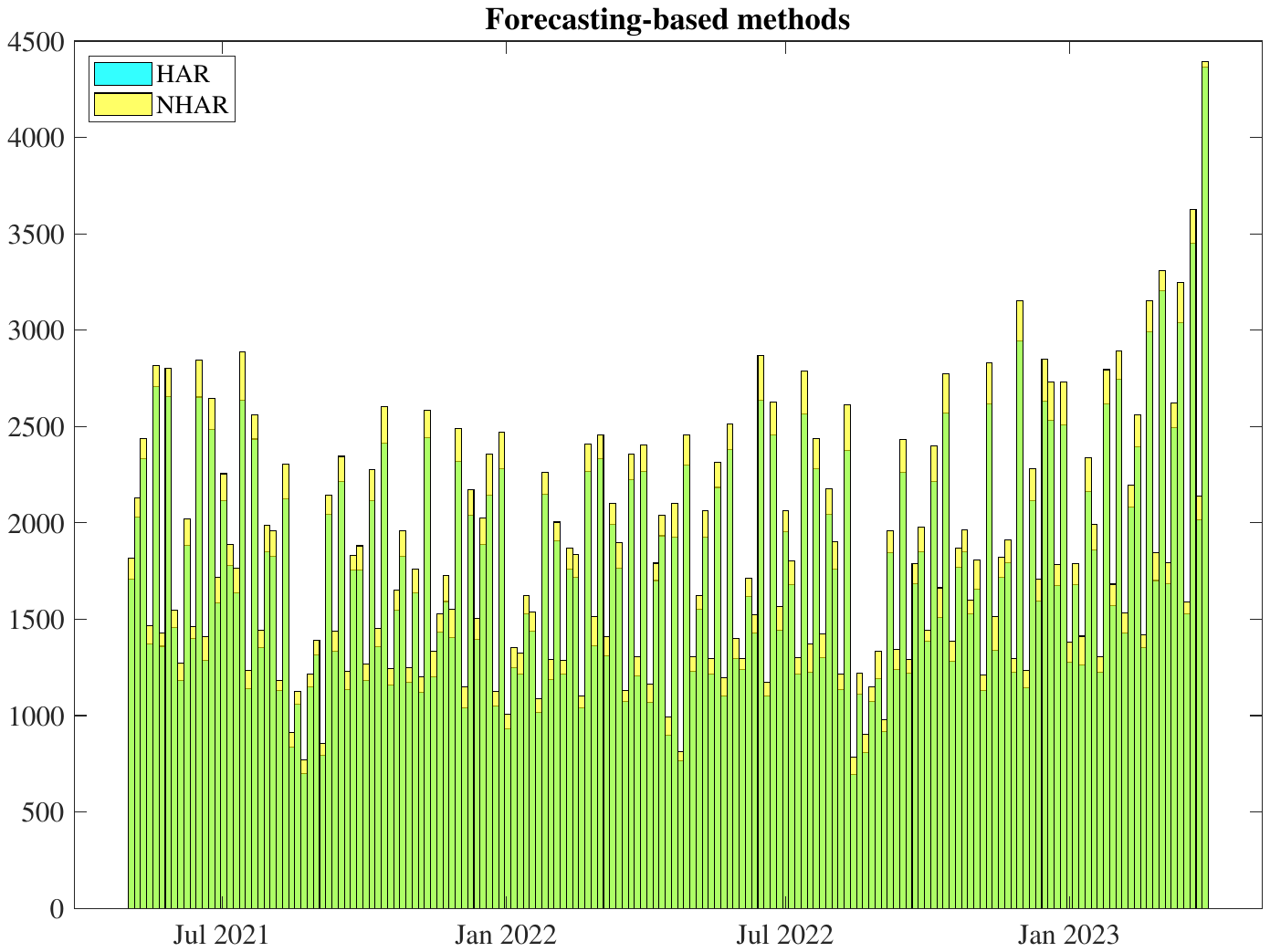}
\caption{\footnotesize Histogram of the total amount of outliersdetected via the two methods.}
\end{subfigure}
\begin{subfigure}[b]{0.45\textwidth}
\includegraphics[width=1.15\linewidth]{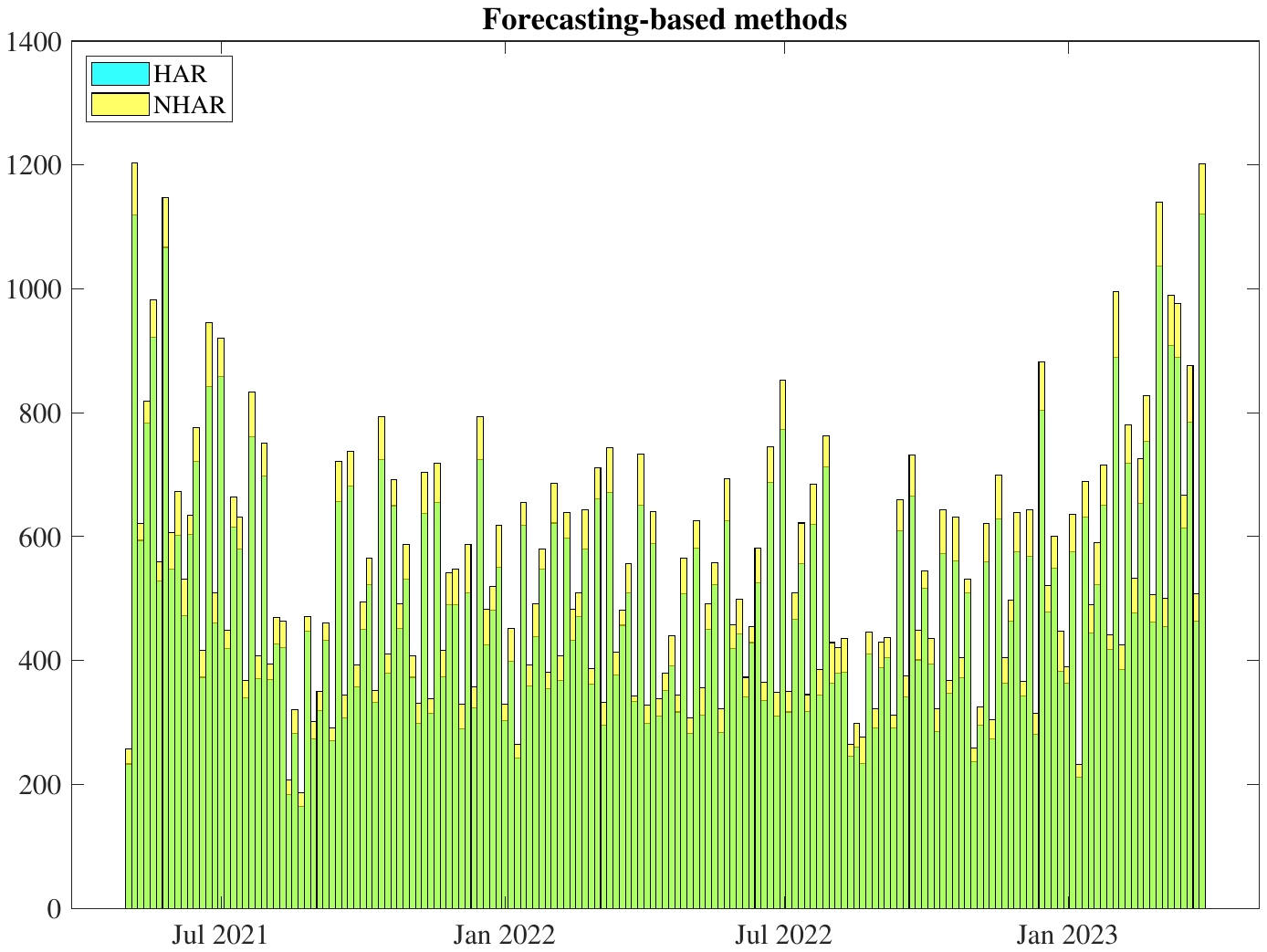}
\caption{\footnotesize Histogram of the negative LSOs detected via the two methods.}
\end{subfigure}
\begin{subfigure}[b]{0.45\textwidth}
\includegraphics[width=1.15\linewidth]{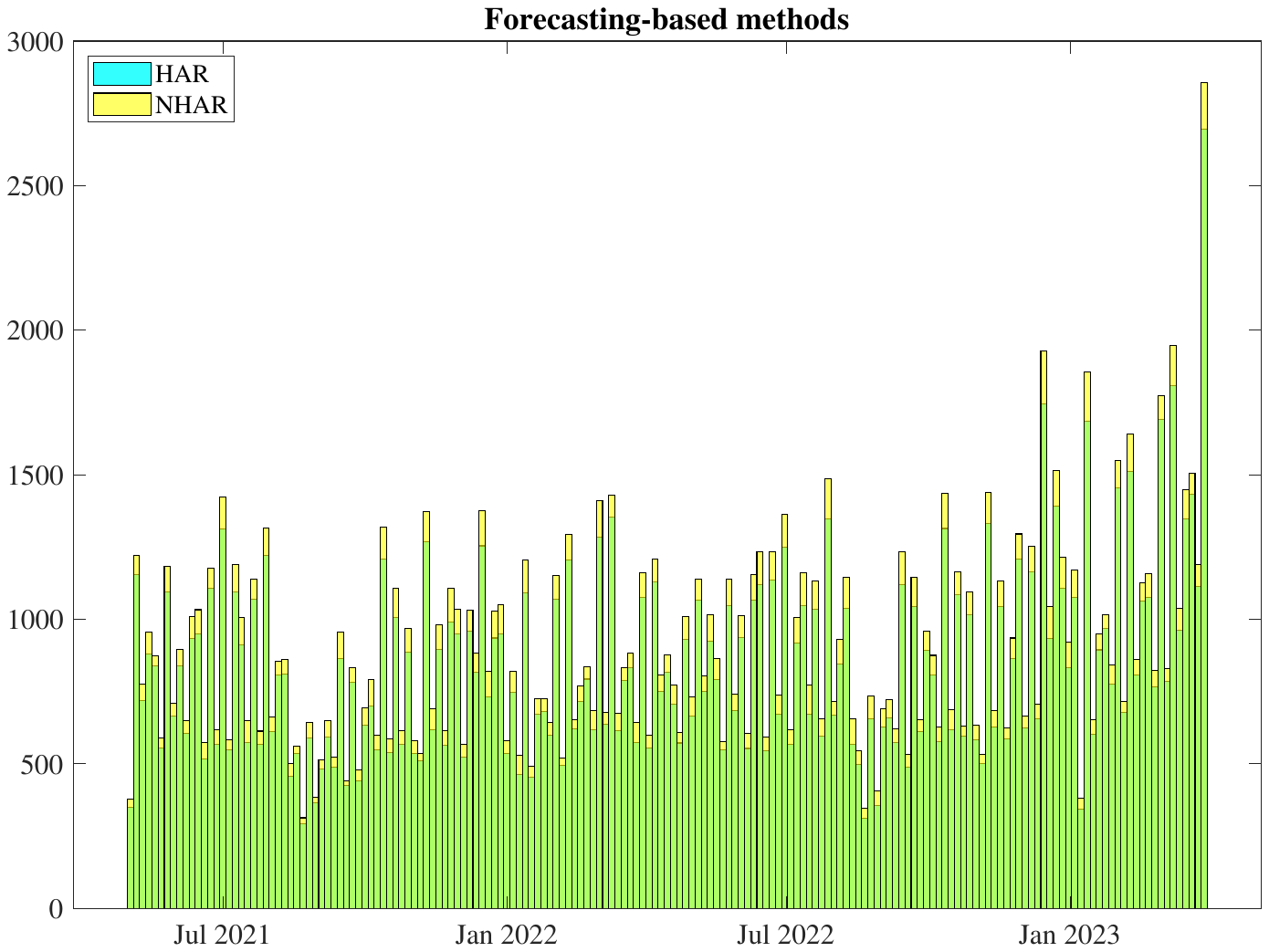}
\caption{\footnotesize Histogram of the positive LSOs detected via the two methods.}
\end{subfigure}
\begin{subfigure}[b]{0.45\textwidth}
\includegraphics[width=1.15\linewidth]{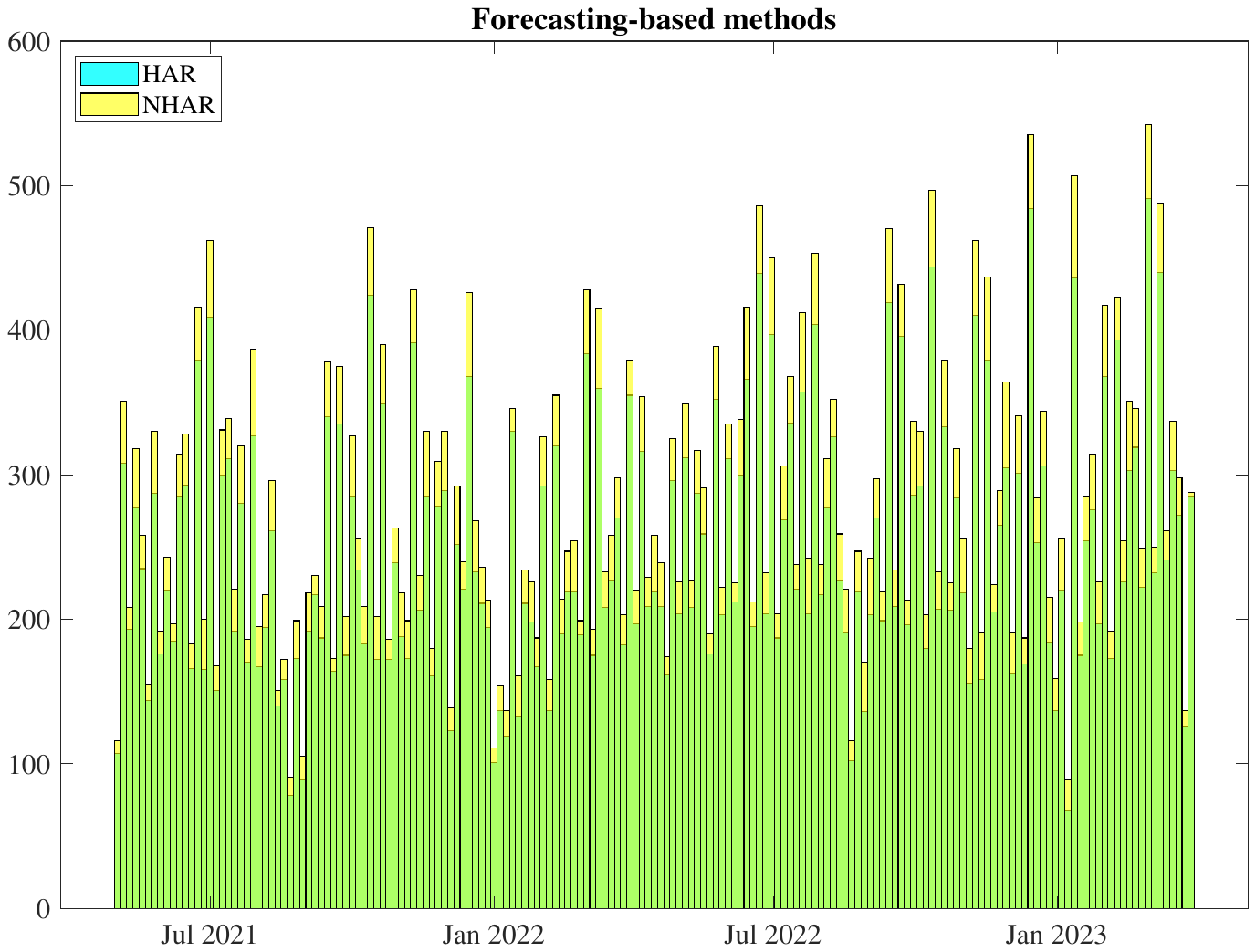}
\caption{\footnotesize Histogram of the AOs detected via the two methods.}
\end{subfigure}
\end{center}
\caption{\footnotesize Distribution of the outliers detected via the RobHAR and RobNHAR methods considering the entire dataset $\bY$.}
\label{hist2M}
\end{figure}

\subsection{Real-time point anomaly detection}

Financial institutions may find it useful to perform point anomaly detection in real time as new information on users' bank account balances becomes available. However, updating the model each day as new data arrive can be computationally expensive. In our framework, we address this issue by performing out-of-sample predictions using the forecasting-based methods applied in the previous subsection.

Suppose the model is fitted up to time $t = T$, so that the robust residuals are available as
\begin{equation}
\hat{r}_{t,i} = y_{t,i} - \bx_{t,i} \bbeta_{i}^{(T)},
\label{RT1}
\end{equation}
for $i = 1, \dots, d$ and $t = 1, \dots, T$, where $\bbeta_{i}^{(T)}$ is estimated via the LTE using the dataset $\lbrace \by_1', \dots, \by_T' \rbrace$.

The out-of-sample prediction of the robust residuals is then computed as
\begin{equation}
\tilde{r}_{T+1,i} = \hat{f}(\hat{r}_{T,i}, \bar{r}_{T-6,i}, \bar{r}_{T-29,i}),
\label{RT2}
\end{equation}
where:
\begin{enumerate}
\item If using the RobHAR method, then $\hat{f}(\hat{r}_{T,i}, \bar{r}_{T-6,i}, \bar{r}_{T-29,i}) = \bv_{T+1}' \hat{\bphi}_i^{(T)}$, where $\hat{\bphi}_i^{(T)} \in 	\mathbb{R}^3$ is estimated via LTE from the robust residuals $\hat{r}_{1,i}, \dots, \hat{r}_{T,i}$ for $i=1,\cdots,d$.
\item If using the RobNHAR model, then $\hat{f}(\hat{r}_{T,i}, \bar{r}_{T-6,i}, \bar{r}_{T-29,i}) = f_{\hat{\bTheta}}(r_{T,i}, \bar{r}_{T-6,i}, \bar{r}_{T-29,i})$, where the feedforward network is trained on 150,000 time series randomly extracted from $\lbrace \br_1, \dots, \br_d \rbrace$.
\end{enumerate}

Once the new observations $y_{T+1,i}$ for $i=1,\cdots,d$ become available, the out-of-sample squared prediction error is computed as:
\begin{equation}
\tilde{\epsilon}_{T+1,i}^2 = (\hat{r}_{T+1,i} - \tilde{r}_{T+1,i})^2,
\label{RT3}
\end{equation}
where $\hat{r}_{T+1,i} = y_{T+1,i} - \bx_{T+1,i} \bbeta_{i}^{(T)}$.

We carry out a forecasting experiment by setting $T = 700$, and performing point anomaly detection iteratively using equations (\ref{RT1})-(\ref{RT3}) on both the robust residuals and their first differences, for $T+1$ to $T+30$, without updating the model at each step.

\begin{figure}
\begin{center}
\includegraphics[width=1.00\linewidth]{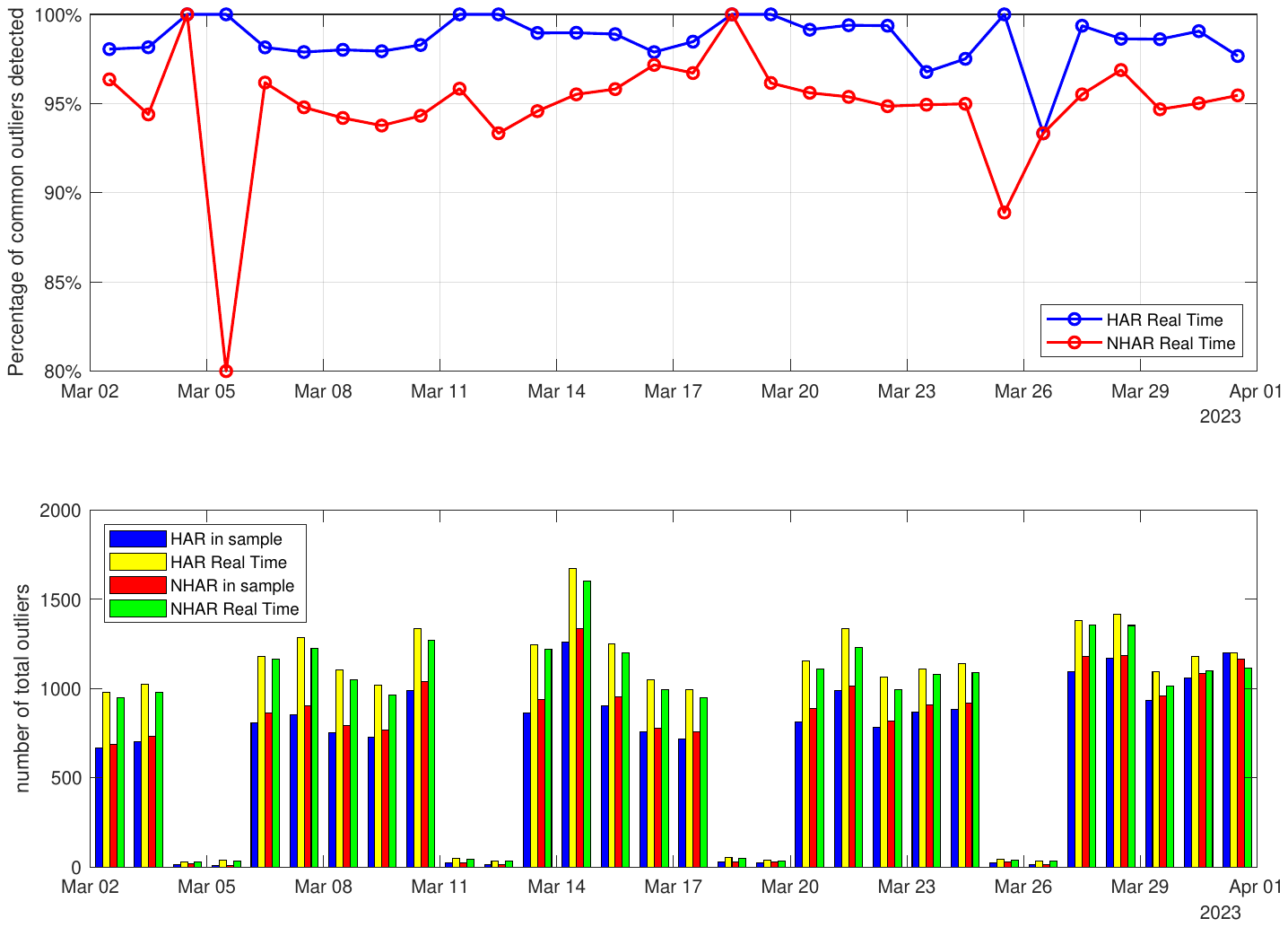}
\end{center}
\caption{\footnotesize Percentage of common outliers detected (on the top) and total number of outliers detected (on the bottom) by the RobHAR and RobNHAR methods, based on real-time out-of-sample and in-sample point anomaly detection. }
\label{figRT}
\end{figure}

The results are reported in Figure~\ref{figRT}. The top panel shows the percentage of outliers detected by the in-sample procedure (carried out in the previous subsection) that are also identified by the real-time out-of-sample method (based on equations (\ref{RT1})-(\ref{RT3})), representing the overlap between in-sample and real-time detections for the two forecasting-based approaches.

The bottom panel reports the total number of outliers detected at each time step by both the HAR and NHAR methods, using both the in-sample and real-time out-of-sample procedures.
Since out-of-sample prediction errors are naturally larger than in-sample ones, real-time anomaly detection identifies a larger number of outliers, particularly with the RobHAR method. The bottom panel also reveals a clear weekly cycle in the anomalies, with most outliers being detected during weekdays.

In summary, both the real-time out-of-sample approaches detect approximately 93-98\% of the in-sample outliers at each time step, with RobHAR performing slightly better than its non-linear counterpart. This suggests that both approaches guarantee satisfactory performance in detecting anomalies in real time, without the need to update the model at each time step.

\section{Clustering the ISP dataset}
\label{secClu}

The methodology applied in the previous sections enables the detection of point anomalies within the ISP dataset.
However, a high percentage of time series (approximately 20-21\%) remain flagged as contaminated, suggesting that the outlier detection process may be overly sensitive.
To further validate these findings and attempt to isolate the genuinely contaminated subset, we perform a cluster analysis on the full dataset.

Clustering is applied to the full sample using the k-means algorithm \shortcite{Lloyd1982}, which minimizes the squared Euclidean distance between observations and their corresponding cluster centroids, with each centroid representing the mean of the points within its cluster.
We use 69 statistical features extracted from the robust residuals as input variables (listed in Table \ref{tfeau}).
Given that k-means relies on a random initialization of the cluster centroids, we perform 50 replications and retain the solution with the lowest total within-cluster variance.

The optimal number of clusters is determined by examining the total within-cluster sum of squared distances as a function of the number of clusters. As the number of clusters increases, this quantity naturally decreases; we select the value of $K$ at which additional clusters lead to only marginal improvements. Based on this criterion, $K = 5$ appears to be a suitable choice.

Table \ref{tCluster} reports the clustering results in the first four columns, which include: the number of time series assigned to each cluster; the average Euclidean distance between a given cluster's centroid and the centroids of the other clusters (inter-cluster distance); and the average distance between each point and its respective cluster centroid (intra-cluster distance).

The last six columns show: (i) the percentage of time series in each cluster that overlap with the set of contaminated series flagged by the RobHAR and RobNHAR procedures (HAROtlrs and NHAROtlrs); (ii) the percentage overlapping with the subset of series affected by LSOs (HARLSOs and NHARLSOs); and (iii) the percentage overlapping with those contaminated by AOs (HARAOs and NHARAOs).

Notably, the 1st cluster exhibits the second highest average inter-cluster distance and the lowest percentage of shared anomalies. In contrast, the 2nd and 4th clusters exhibit the highest overlap with the contaminated series, with relatively large inter-cluster distances from the 1st cluster (17.80 and 20.54, respectively). These findings suggest that the k-means algorithm is able to separate regular from anomalous series to some extent, and that the 1st cluster likely represents a "clean" or safe group of users.

The 4th cluster, in particular, is the least compact (with an average intra-cluster distance of 15.23), and the smallest, accounting for only 1.51\% of the full dataset. Additionally, it present the highest average inter-cluster distance and overlaps with over 50\% of the total anomalies detected by the RobHAR and RobNHAR methods, making it a strong candidate for capturing truly anomalous behavior.
To refine our classification, we define a subset of the 4th cluster composed of time series that are also flagged as contaminated by RobHAR and RobNHAR. This subset, which represents approximately 0.8\% of the entire dataset, is identified as potentially contaminated by outliers.
The identification of an "anomalous" cluster can also be valuable for real-time anomaly detection. For instance, it enables targeted forecasting by applying a pre-fitted model specifically to the time series belonging to this cluster. As new data points become available, predictions can be updated in real time, focusing exclusively on the potentially contaminated series.

Finally, it is worth noting that the proportion of shared AOs is relatively low in the 2nd and 4th clusters and comparatively high in the 1st cluster, while the opposite trend holds for LSOs. This pattern suggests that the clustering algorithm is less effective at distinguishing between different types of anomalies, i.e., AOs versus LSOs.

\begin{table}
\centering
\resizebox{0.95\textwidth}{!}{
\begin{tabular}{l|c}
\hline
\hline
 Mean &  $\frac{1}{n}\sum_{t=1}^n \hat{r}_t$  \\
  Standard Deviation &  $\sqrt{\frac{1}{n-1}\sum_{t=1}^n (\hat{r}_t-\bar{r})^2}$  \\
   Maximum &  $\mbox{max}(\hat{r}_t)$  \\
 Minimum &  $\mbox{min}(\hat{r}_t)$  \\
 Median Absolute deviation    & $\mbox{med}( \vert \hat{r}_t - \mbox{med}(\hat{r}_t) \vert)$    \\
 Interquartile range & $Q_r(0.75)-Q_r(0.25)$  \\
 Skewness  &   $\frac{1}{(n-1)\hat{\sigma}^3}\sum_{t=1}^n (\hat{r}_t-\bar{r})^3$        \\
 Kurtosis  &     $\frac{1}{(n-1)\hat{\sigma}^4}\sum_{t=1}^n (\hat{r}_t-\bar{r})^4$      \\
  Robust Trend Intercept & $ \hat{\beta}_{\nu,0}$    \\
   Robust Linear Trend Coefficient & $ \hat{\beta}_{\nu,1}$    \\
      Robust Quadratic Trend Coefficient & $ \hat{\beta}_{\nu,2}$     \\
            Robust Amplitude of the 1st and 2nd harmonics & $ \sqrt{ \hat{\beta}^2_{\gamma,j} +  \hat{\beta}^{*2}_{\gamma,j}}$    \\
             Robust Phase of the 1st and 2nd harmonics & $ \arctan \biggr(\frac{\hat{\beta}_{\gamma,j}}{ \hat{\beta}^{*}_{\gamma,j}} \biggr)$    \\
              First 50 coefficients of the autocorrelation function & $(n-j)^{-1}\sum^n_{t=1}\hat{r}_t \hat{r}_{t-j} \>\>j=1,\cdots,50$ \\
                 Periodogram maxima close to the zero, weekly and monthly freq.&   $I_r(0),I_r(2 \pi/7), I_r(2\pi/30)$ \\
            \hline
             \hline
\end{tabular}}
\caption{\footnotesize Statistical features introduced as an input in the k-means algorithm to cluster the ISP time series.}
\label{tfeau}
\end{table}

\begin{table}
\centering
\resizebox{1.00\textwidth}{!}{
\begin{tabular}{c||rrrr|ccc|ccc}
\footnotesize{Cluster}  &\footnotesize{NumTS} & \footnotesize{PercTS} & \footnotesize{InterClsDis}  &  \footnotesize{IntraClsDis}   &  \footnotesize{HAROtlrs}   &  \footnotesize{HARLSOs}  & \footnotesize{HARAOs} & \footnotesize{NHAROtlrs}   &  \footnotesize{NHARLSOs}  & \footnotesize{NHARAOs} \\
\hline
\hline
1 & 296,352  &  19.71\% &  14.1799 &     4.3438  &8.51\%  &  2.08\% & 4.91\% & 7.91\%  &  2.03\% & 5.31\%\\
2 & 375,038  &  24.94\% &  12.0043 &  3.7648   &   33.42\%  &  29.39\%  & 0.53\% &36.09\%  &  31.57\% & 0.67\%\\
3 & 393,724  &   26.18\%&  10.3098 &  3.8600 &   14.32\%  &    8.87\% & 4.59\% &15.19\%  &  9.66\% & 5.10\%\\
4 & 22,646  &  1.51\% &  15.2267   &  14.9607 &  51.92\%   & 47.02 \% & 1.70\% &56.08\%  & 50.41\% & 2.18\%\\
5 & 416,120  &  27.67\%  & 9.2991 & 3.4916 &  19.98\%  &  17.03\% &  1.96\% &21.98\%  & 19.07\% & 2.28\%\\
\hline
\hline
\end{tabular}}

\caption{\footnotesize Clustering results are reported in terms of the number of time series per cluster (NumTS), the percentage of time series in each cluster (PercTS), the average Euclidean distance between the centroids of the $K$ clusters (InterClsDis), and the average Euclidean distance between all points within a cluster and their respective centroid (IntraClsDis).
The last three columns report the percentage of shared anomalies between the series in each cluster and the number of series flagged as contaminated  by the RobHAR and RobNHAR methodologies (HAROtlrs and NHAROtlrs), as well as the series contaminated by the LSOs (HARLSOs and NHARLSOs)and AOs (HARAOs and NHARAOs) detected via the procedure described in section \ref{cusum}.
}
\label{tCluster}
\end{table}

\section{Conclusions}

In this paper, we have performed point anomalies detection on the ISP dataset by analyzing the entire sample, comprising approximately 2.6 million time series, as well as two sub-samples consisting of 50,000 and 5,000 time series, respectively.

Distance-based methods such as the OGK, MRCD, and COM robust covariance estimators can be applied to the low-dimensional datasets; however, these techniques become infeasible when dealing with the full sample. In such cases, a forecasting-based strategy like the RobHAR and RobNHAR methodologies may be preferable. In the analysis of the low-dimensional datasets, the different approaches yield nearly identical results. The distance-based methods capture over 90\% of the outliers detected by the forecasting-based approach, and the percentage of common anomalies among the distance-based methods exceeds approximately 96\%. These results suggest that forecasting-based methods, although they process each series individually, can be a valuable alternative to distance-based methodologies, which account for cross-series covariances when performing point anomaly detection.

In the end, our methodology flags approximately 3\% of all daily user transactions as potential anomalies, identifying the time of occurrence for each outlier. It is important to emphasize that this does not necessarily indicate malicious activity. Indeed, the robust methods employed in this study may flag as outliers exceptional transactions that simply deviate from a user's typical behavior (e.g., unexpected large cash withdrawals, incoming wire transfers to settle old debts, or large deposits and withdrawals occurring on the same day).

Therefore, the presence of a point anomaly should be verified in a subsequent step, where additional contextual information can be checked, such as the time of the transaction, the recipient's country, whether the recipient is a known contact, and so on.

Similarly, analyzing the distribution of negative LSOs can help identify whether potentially suspicious activity is affecting multiple users during specific times of the year when large withdrawals are less likely to be justified (e.g., outside typical holiday periods). Analogous findings emerge from the distribution of AOs.

The set of potentially contaminated series can be further refined through cluster analysis, by identifying time series that are both flagged as contaminated by the robust methodology and belong to a small, isolated cluster characterized by high internal variability. According to this criterion, approximately 0.8\% of the entire sample (12,031 users accounts) can be flagged as potentially contaminated. Moreover, identifying such an "anomalous" cluster can support real-time anomaly detection, allowing predictions to be performed exclusively on the series within the contaminated cluster using a pre-trained model, without the need to re-process the entire dataset.

\section*{Acknowledgement}

This paper is supported by Italian Research Center on High Performance Computing Big Data and Quantum Computing (ICSC), project funded by European Union - NextGenerationEU - and National Recovery and Resilience Plan (NRRP) - Mission 4 Component 2 within the activities of Spoke 3 (Astrophysics and Cosmos Observations).

\section*{Code availability}

The source code and scripts used for the analyses in this paper are openly available at the GitHub repository: 
\href{https://github.com/ICSC-Spoke3/ATS}{\texttt{https://github.com/ICSC-Spoke3/ATS}}.

\bibliographystyle{apalike}
\bibliography{References}

\appendix

\section{Least trimmed Estimator (LTE)}
\label{A1}

In the following, we describe step by step the procedure to obtain the trimmed estimator $\hat{\bbeta}=(\hat{\bbeta}_\nu',\hat{\bbeta}_\gamma')'$ for equation (\ref{detTrend}).
Assume a quadratic trend and a two harmonics deterministic cycle, such that $v=c=2$.
Let $\bx_{reg,t}=(1, t, t^2, \cos(\lambda t), \sin(\lambda t), \cos(2\lambda  t),\sin( 2\lambda  t))$, and set a preliminary value $SS^*$ to a very large number (e.g. $10^{20}$).

\begin{enumerate}

\item Draw a random sample $\bs$ of size $(1+v+2c)$ from the integers ${1, \dots, n}$.

\item Given the indexes $\bs$, compute $\bbeta_\bs=(\bX_{reg,\bs}'\bX_{reg,\bs})^{-1}\bX_{reg,\bs}'\bY_{\bs}$.

\item Compute the residuals $\tilde{\br}_t=\bY-\bX_{reg}\bbeta_\bs$.

\item Order the $h$ smallest residuals to obtain the set $\bi = i_1,i_2, \cdots,i_h$, with $\tilde{r}_{i_1} \leq \tilde{r}_{i_2} \leq \cdots \leq \tilde{r}_{i_h}$.

\item Compute the estimator $$\bbeta^{(j)}=(\bX_{reg,\bi}'\bX_{reg,\bi})^{-1}\bX_{reg,\bi}'\bY_{\bi}$$ with corresponding residuals $\tilde{\br}_t=\bY-\bX_{reg}\bbeta^{(j)}$.

\item Obtain the sum of squared residuals $SS^{(j)}=\sum^h_{t=1} r^{2}_t$

\item If $SS^{(j)}<SS^{*}$, update $\hat{\bbeta}=\bbeta^{(j)}$ and set $SS^{*} =SS^{(j)} $.
\end{enumerate}

Repeat the above steps for \( j = 1, 2, \dots, 500 \) to obtain the final trimmed estimator $\hat{\bbeta}$.


%
%
%
%
%
%
%
%
%

\end{document}